\documentclass{openmoss}

\usepackage{array}
\newcolumntype{L}[1]{>{\raggedright\arraybackslash}p{#1}}
\newcolumntype{R}[1]{>{\raggedleft\arraybackslash}p{#1}}
\newcolumntype{C}[1]{>{\centering\arraybackslash}p{#1}}

\graphicspath{{assets/}{figures/}{./}}

\newcommand{\benchmarkname}{ContextWeave}

\renewcommand{\headerlogo}{%
  \includegraphics[height=9mm]{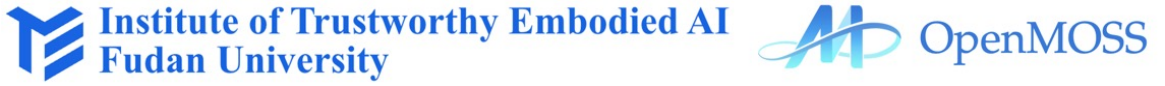}%
}
\newcommand{\bytedancelogo}{\includegraphics[height=9mm]{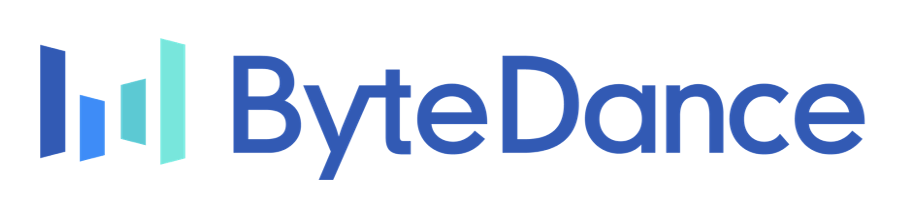}}
\fancypagestyle{firststyle}{%
  \fancyhead[L]{\vskip 6mm \headerlogo}%
  \fancyhead[R]{\vskip 6mm \bytedancelogo}%
}
\title{ContextWeave: A Real-World Workflow Benchmark}

\author[1,*]{Bo Wang}
\author[1,2,*]{Yuqian Yao}
\author[1,*]{Enxi Wang}
\author[1,2,*]{Luozhijie Jin}
\author[1,2,*]{Yang Liu}
\author[1,2]{\\Yiran Suo}
\author{Yuxuan Cai}
\author[1]{Enyu Zhou}
\author[1]{Yufei Gao}
\author[1]{Honglin Guo}
\author[1,2]{Tianyu Huai}
\author[1,2]{Li Ji}
\author{Zhikai Lei}
\author[1]{Bufan Li}
\author{Lizhi Lin}
\author{Jinxiu Liu}
\author[1]{Jie Yang}
\author[1]{Jiazheng Zhou}
\author[1,2]{Maosen Zhou}
\author[1,2]{Pengfang Qian}
\author[1]{Shichun Liu}
\author[3]{Guanshan Liu}
\author[3]{Hao Zheng}
\author[3]{Yunhao Yu}
\author{Hang Yan}
\author[3]{\\Jihua Kang}
\author[1,\dagger]{Xinchi Chen}
\author[1,2,\dagger]{Xipeng Qiu}

\affil[1]{Fudan University}
\affil[2]{Shanghai Innovation Institute}
\affil[3]{ByteDance}

\abstract{Memory is essential as language agents move from isolated tasks to long-horizon, stateful workflows, yet existing evaluations often reduce it to retrieval or question answering. We introduce \benchmarkname, a longitudinal benchmark that evaluates whether recalled experience improves downstream agent performance in realistic office-work streams. \benchmarkname{} reconstructs privacy-preserved, multi-month workflows of 14 participants into 1,005 executable tasks, including 568 core evaluation tasks, with instructions, containerized environments, trajectories, and task-specific rubrics. It measures workspace quality and alignment with participant-specific preferences, complemented by diagnostics of relevance, continuity, solvability, and robustness to misleading recall. Across six memory components under a fixed model, the strongest configuration raises Workspace Score from 68.08 to 78.20 and Preference Score from 41.50 to 70.60. With a fixed memory component, recall improves both outcomes for all five tested base models, although gains vary substantially. Our analysis shows that actionable, experience-rich memory supports workflow continuation and reduces redundant exploration more effectively than compact summaries, while it can also be more susceptible to misleading recall. These findings motivate memory systems that optimize not only retrieval relevance but also reliable use during execution.
}

\checkdata[GitHub Repo]{\href{https://github.com/OpenMOSS/ContextWeave}{github.com/OpenMOSS/ContextWeave}}
\checkdata[Lead contact]{\href{mailto:oliver.b.wang2001@gmail.com}{oliver.b.wang2001@gmail.com}}
\checkdata[Correspondence]{\href{mailto:xc_chen@fudan.edu.cn}{xc\_chen@fudan.edu.cn}; \href{mailto:xpqiu@fudan.edu.cn}{xpqiu@fudan.edu.cn}}

\newtcolorbox{promptbox}[2][]{
  colback=white,
  coltext=black,
  arc=3mm,
  boxrule=0.5pt,
  colframe=black!60!white,
  title={#2},
  colbacktitle=black,
  coltitle=white,
  fonttitle=\bfseries,
  top=8pt,
  bottom=8pt,
  left=10pt,
  right=10pt,
  breakable,
  before upper={
    \linespread{1}\selectfont
    \setlength{\parskip}{1ex plus 0.2ex minus 0.2ex}
    \setlength{\parindent}{0pt}
  },
  #1
}

\begin{document}
\maketitle
\authornote{$^{*}$Equal contribution.\quad$^{\dagger}$Corresponding authors.}

\section{Introduction}
Memory is increasingly recognized as a central capability for language agents as they move beyond isolated prompt-based problem solving toward long-horizon, stateful workflows. Early work primarily focused on self-contained mathematical reasoning and short-form code generation tasks \citep{lewkowycz2022minerva,roziere2023codellama,
azerbayev2024llemma,shao2024deepseekmath}. Modern agents are increasingly expected to handle complex workflows that unfold over extended trajectories \citep{mialon2023gaia,zhou2024webarena,jimenez2024swebench}. This creates a fundamental tension between the bounded context window of current models and the growing demands of long-horizon agentic workflows, making memory essential for retaining task-relevant information, reusing accumulated experience across sessions, and improving through repeated interaction.

Building effective memory systems requires evaluation protocols that faithfully measure whether memory improves agent behavior. Existing evaluations capture different fragments of this problem. Long-context, retrieval-oriented, and conversational memory benchmarks typically evaluate whether models can retrieve, summarize, or reason over information embedded in long inputs or multi-session dialogues \citep{hsieh2024ruler,bai2024longbench,maharana2024locomo,wu2025longmemeval}. While useful, these settings largely reduce memory to context-based question answering. End-to-end agent benchmarks can reflect the contribution of memory components within a full agent system, but they are usually organized as independent task instances and thus provide limited evidence about cross-episode memory. Self-improvement methods \citep{shinn2023reflexion,zhao2024expel} evaluate evolution through repeated trials or epochs on existing task distributions. However, this repeated-trial setting misses the sparse, implicit, and user-specific dependencies that characterize longitudinal real-world use.


To address this gap, we first ground memory evaluation in its connection to learning. In neuroscience, learning and memory are tightly coupled processes: organisms acquire experience from the environment, retain it, and use it to shape future behavior \citep{okano2000learning, kandel2014molecular}. This view suggests a simple operational criterion for agent memory: a useful memory system should improve downstream task performance as the task stream progresses. Motivated by this criterion, we introduce \benchmarkname, a longitudinal benchmark for evaluating memory systems in realistic office-work scenarios. Built from real users' work documents collected over multiple months, and processed through strict de-identification, data cleaning, and human annotation, each benchmark instance contains an ordered user-specific task stream with reconstructed instructions, execution trajectories, intermediate artifacts, and evaluation rubrics. At test time, we compare an agent's performance on a target task with and without access to preceding task histories, providing a controlled and reproducible measure of memory-induced improvement in realistic cross-session workflows.

Our experiments reveal that useful memory is better understood as support for workflow continuation than as retrieval in isolation. Without recall, agents often produce superficially complete outputs that fail to preserve latent workspace state and participant-specific work practices; access to prior experience improves both Workspace and Preference scores for every tested base model. Yet relevance alone is insufficient: in our tested configurations, memory closer to concrete in-context experience provides more actionable evidence and reduces repeated exploration more effectively than compact summaries, but is also more susceptible to errors caused by misleading recall. These results motivate evaluating memory through downstream outcomes, behavioral effects, and robustness together.

Our contributions are as follows:
\begin{itemize}
    \item \textbf{A longitudinal benchmark grounded in real workflows.}
    We introduce \benchmarkname, which contains 1,005 reconstructed executable tasks from the multi-month workflows of 14 participants, including 568 primary-workflow tasks used for evaluation. Each participant-specific stream preserves privacy-protected histories, temporal dependencies, executable environments, and task-level evaluation criteria.

    \item \textbf{An executable and controlled evaluation protocol.}
    We develop a pipeline that reconstructs instructions and missing observations from raw worklogs, verifies each task through execution, and prevents accumulated execution drift through controlled trajectory alignment. The protocol measures downstream Workspace and Preference outcomes while diagnosing Relevance, Continuity, Solvability, and Hallucination Robustness.

    \item \textbf{A systematic empirical study of agent memory.}
    We evaluate six memory components and five base models under controlled with- and without-recall settings. The results show that memory improves downstream outcomes across all tested models, while its benefits and risks depend on the actionability of recalled experience and the model's ability to apply it reliably.
\end{itemize}

\section{Related Work}
Memory has been studied under different contexts in language-model and agent research, each implying a different evaluation target. In long-context modeling, memory emphasizes information access over extended inputs; in agent systems, it often refers to working memory within an episode or episodic memory across sessions. We therefore organize existing memory evaluations into four broad categories and discuss them in turn.

\subsection{Long History QA}
A common line of work evaluates memory through a query-over-history format: a model is given a long context or conversation history and asked to answer a query grounded in that record. RULER extends Needle-in-a-Haystack-style evaluation with retrieval, multi-hop tracing, and aggregation tasks over synthetic long contexts \citep{hsieh2024ruler}. This form of evaluation is widely used in long-context modeling \citep{fang2025artificial,bai2024longbench}. Long-term conversational benchmarks such as LoCoMo and LongMemEval instantiate a similar format over multi-session dialogue histories, with questions targeting facts, events, preferences, temporal relations, or updated user information \citep{maharana2024locomo,wu2025longmemeval}. These benchmarks capture an important aspect of memory: whether key information can be extracted from historical context and used when queried. However, because they reduce evaluation to answering a query over a given record, they leave open whether memory improves downstream performance on more complex agentic tasks.

\subsection{Independent Agent Task}

Another line of work studies memory through end-to-end agent tasks. Benchmarks such as GAIA, SWE-bench, and DeepSWE require agents to solve complex problems over extended trajectories, thereby implicitly testing their ability to maintain, organize, and use task state during execution \citep{mialon2023gaia,jimenez2024swebench,huang2026deepswe}. These benchmarks provide useful probes of working memory in realistic agent tasks, even though they are not designed specifically for memory. To approximate episodic memory, prior work often runs agents for multiple trials or epochs on tasks such as HotpotQA and ALFWorld, using changes in end-to-end performance as evidence of experience reuse \citep{yang2018hotpotqa,shridhar2021alfworld,shinn2023reflexion,zhou2025memento,zhou2026mementoskills}. This extends memory evaluation beyond QA-style interfaces by measuring how accumulated experience affects agent behavior in task execution. However, real longitudinal use involves dependencies that may be sparse, implicit, and user-specific, making it unclear whether repeated-trial evaluations capture memory effectiveness in realistic workflows.

\subsection{Constructed multi-session agent tasks}

Recent benchmarks have begun to evaluate memory in multi-session agentic settings. MemoryArena constructs human-crafted tasks with explicitly interdependent subtasks, where agents must distill experience from earlier sessions and use it to guide later actions in a Memory-Agent-Environment loop \citep{he2026memoryarena}. This direction moves beyond isolated recall and evaluates memory through downstream agent behavior. However, the cross-session dependencies in these benchmarks are manually constructed, which may not fully reflect the sparse and implicit dependencies that emerge in real-world workflows.

\subsection{Component-level memory diagnostics}

Some benchmarks evaluate memory by first specifying the functional capabilities that a memory system is expected to support, and then designing targeted metrics for each capability. For example, MemBench measures LLM-agent memory across factual and reflective memory, participation and observation scenarios, and metrics such as effectiveness, efficiency, and capacity \citep{tan2025membench}. MemoryAgentBench decomposes memory agents into competencies including accurate retrieval, test-time learning, long-range understanding, and selective forgetting \citep{hu2025memoryagentbench}. MemoryBench simulates user feedback to study memory and continual learning in LLM systems, covering declarative and procedural memory needs \citep{ai2025memorybench}. These benchmarks are useful for diagnosing whether a system can store, retrieve, update, abstract, or forget information properly. Such component-level measurements are better viewed as explanatory signals than as direct estimates of downstream agent performance: a memory system that performs well on retrieval or update metrics may not necessarily help an agent select and use past experience effectively in future workflows.

\subsection{Memory Components in LLM Agents}
Beyond memory evaluation, we briefly review how memory components are constructed in LLM-based agents. Existing systems differ in how they abstract, organize, and update past experience. MemoryBank maintains long-term user memories through continuous updates and an Ebbinghaus-inspired forgetting and reinforcement mechanism, enabling agents to adapt to users over time \citep{zhong2024memorybank}. Mem0 dynamically extracts, consolidates, and retrieves salient information from conversations, with a graph-based variant for capturing relational structure \citep{mem02025}. LangMem is a software framework that provides modular utilities for constructing semantic, episodic, and procedural memories, supporting both in-conversation and background memory updates \citep{langmem2025}. A-MEM organizes experience as Zettelkasten-inspired structured notes and dynamically creates and updates links among memory items \citep{xu2025amem}. MemOS treats memory as a manageable system resource and unifies the representation, scheduling, and evolution of plaintext, activation-based, and parameter-level memories \citep{li2025memos}. Together, these systems span a design space of memory abstraction, organization, and lifecycle control: compact fact- or summary-based stores favor efficient retrieval, linked-note systems maintain explicit relationships among memories, and OS-style frameworks coordinate heterogeneous memory representations.

\section{Design Principle}
\subsection{Mathematical Formulation}
In real-world scenarios, agents often interact with users over a sequence of related tasks rather than solving isolated tasks independently. We formulate such interactions as a sequential task stream:
\begin{equation}
D = (T_1, T_2, \dots, T_n), \quad T_1 \rightarrow T_2 \rightarrow \dots \rightarrow T_n ,
\end{equation}
where each $T_i$ represents an individual task encountered by the agent, and the ordering reflects the temporal dependency among tasks.

Given a memory component $M$, we evaluate whether leveraging previous task experiences improves the agent's performance on subsequent tasks. Specifically, for each target task $T_i$, we first evaluate an agent without access to previous experiences, where the performance is denoted as $R(T_i)$.
Then, equipped with a memory component $M$, the agent first processes historical task trajectories:
\begin{equation}
\{T_1, T_2, \dots, T_{i-1}\},
\end{equation}
and obtains a memory representation that is provided when solving the target task $T_i$. The resulting performance is denoted as $R_M(T_i)$.
The contribution of the memory component is measured by the performance gain brought by historical experience:
\begin{equation}
\Delta R_M(T_i) = R_M(T_i) - R(T_i).
\end{equation}
A positive $\Delta R_M(T_i)$ indicates that the memory component successfully enables the agent to leverage previous experience for improving future task execution.

\subsection{Benchmark Design Considerations}
To ensure that the benchmark faithfully reflects memory usage in realistic agent scenarios, we consider four design principles.

\paragraph{Privacy-preserving realism.}

Realistic memory evaluation requires authentic workflow histories rather than artificially constructed interactions. We therefore build our benchmark from real-world collaborative workflows and apply privacy-preserving processing to remove sensitive information while maintaining the structural properties of the original tasks, including temporal order, entity consistency, and task dependencies. Section \ref{subsec:privacy} will discuss this question in detail.

\paragraph{Executable and reproducible environments.}

Memory evaluation should go beyond passive information retrieval and measure whether agents can leverage historical experience during task execution. We therefore provide executable environments that allow agents to directly interact with task contexts. At the same time, we ensure reproducibility by controlling external dependencies and reconstructing unavailable resources when necessary, enabling consistent evaluation across different memory systems. 

\paragraph{Controlled longitudinal evaluation.}

Although real workflows naturally evolve through sequential task execution, directly replaying the entire process introduces efficiency challenges and error accumulation. We therefore decouple historical experience from environment evolution by preserving controlled task trajectories as memory inputs, allowing memory components to process realistic past experiences while maintaining a stable evaluation setting.

\paragraph{Downstream-oriented evaluation.}

The ultimate goal of memory is to improve an agent's ability to complete future tasks, rather than merely optimize intermediate memory operations. Therefore, we prioritize downstream task outcomes, including task execution quality and alignment with user preferences, while treating memory-specific measurements as complementary diagnostics for understanding agent behavior.

\section{Data Construction}
We construct our benchmark from longitudinal records of real-world workflows rather than manually designed task sequences. Specifically, we collect complete work histories from an open-source project involving 14 participants over multiple months.

Starting from these raw workflow records, we construct executable agent tasks through a series of processing steps, including task stream extraction, privacy-preserving data processing, task reconstruction, controlled trajectory alignment, and evaluation design. The resulting benchmark preserves the temporal structure and dependencies of real workflows while transforming human activity traces into reproducible agent task sequences with executable environments and downstream-oriented evaluation criteria.

\begin{figure}[t]
    \centering
    \includegraphics[width=\linewidth]{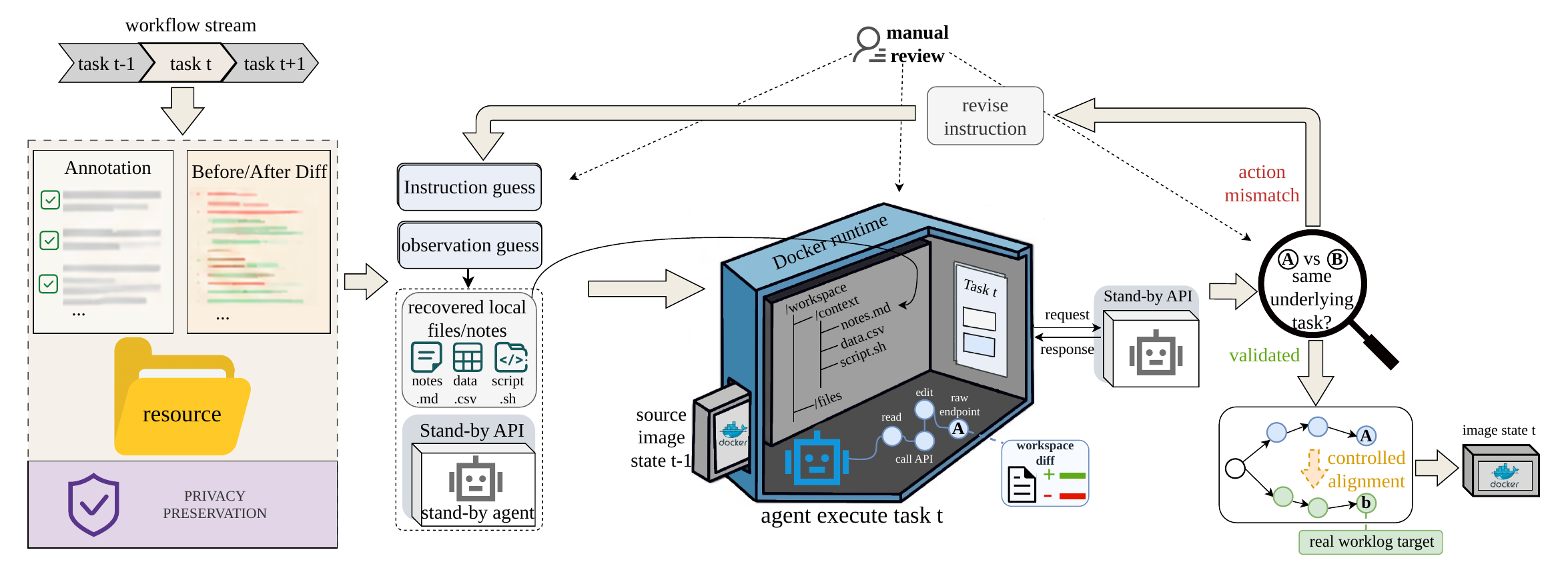}
    \caption{Overview of the benchmark construction pipeline. Privacy-preserved workflow annotations, before/after diffs, and available resources are used to infer instructions and reconstruct task observations as local artifacts or controlled stand-by APIs; each task is then executed in an isolated Docker environment, iteratively reviewed against the original worklog, and aligned to its recorded outcome to provide a canonical state for subsequent tasks.}
    \label{fig:data-construction}
\end{figure}

\subsection{From Real Workflows to Task Streams}
\label{subsec:task-segmentation}
The first step in evaluating episodic memory is to recover realistic task sequences from raw workflow traces. For each participant, we collect historical document-editing events, order them chronologically, and retain the document identifier, local content before and after each edit, and the corresponding document diff. We then use LLM-assisted segmentation to group consecutive events into semantically coherent task units, where each unit corresponds to a user-level objective that could naturally be delegated to an agent. Since real document histories contain noise and interleaved workstreams, we further conduct human verification to refine task boundaries and annotate task-level metadata, including the task objective, whether it belongs to the participant's primary workflow, required external resources, and constraints for later task reconstruction. After human verification, we obtain 1005 task instances in total, including 568 primary-workflow tasks. Detailed annotation guidelines and LLM prompts are provided in Appendix~\ref{sec:app_prompt}.

\subsection{Privacy Preservation}
\label{subsec:privacy}
Real-world workflows contain sensitive information, including personal identifiers, private documents, internal services, local paths, and credentials. To preserve privacy while maintaining task structure, we apply a structure-preserving anonymization process that combines rule-based detection with agent-assisted verification. Specifically, sensitive entities are consistently replaced across the entire task sequence, private domains and paths are normalized, and credential-like information is replaced with format-preserving placeholders. Public resources and non-sensitive information are retained when possible. This procedure preserves entity relationships, structural patterns, and task dependencies without exposing private content.

\subsection{Task Reconstruction}
\label{subsec:task_reconstruction}

Through the steps described in Section~\ref{subsec:task-segmentation}, we transform raw workflow records into segmented worklog entries, each corresponding to a user-level task that could in principle be delegated to an agent. However, these entries primarily capture what changed after a human completed the task, rather than the explicit instruction and execution context required for an agent to perform it. We therefore reconstruct each segmented entry into an executable agent task instance. Specifically, using the worklog evidence and human-annotated metadata obtained in the previous step, we independently construct each instance through three stages: instruction generation, observation and environment construction, and execution-based verification. We describe these stages below.

\paragraph{Instruction generation.}
For each task represented by a segmented worklog entry, we use the LLM to reconstruct its underlying work objective. The model considers the evidence associated with the entry—including the pre-task and post-task document states and the document diff—together with task-level annotations and any available historical context. It is prompted to infer a plausible user request that could have produced the observed outcome and to express that request as a natural agent instruction. The resulting instruction specifies the task goal, necessary resources, expected outcome, and key constraints, while remaining goal-oriented rather than prescribing a step-by-step procedure. This formulation allows the agent to plan its own execution while reducing the risk of leaking the final document content. The prompt used for this request-inference and instruction-generation process is provided in Appendix~\ref{sec:app_prompt}.

\paragraph{Observation and environment construction.}
Real worklogs often omit observations that were available to the human worker but are not preserved in the document history. These missing observations may include:
\begin{itemize}
\vspace{-2pt}
    \item meeting discussions or informal decisions;
    \item local data files or intermediate artifacts;
    \item experiment outputs or model evaluation results;
    \item third-party services, APIs, or tool responses;
    \item results that depend on unstable tool versions or expensive resources such as GPUs.
\vspace{-2pt}
\end{itemize}

We construct task observations using one of two mechanisms, depending on whether the underlying resources can be faithfully, safely, and practically reproduced. When a resource is static and recoverable with high confidence from the workflow evidence, we materialize the relevant information in the container environment as local files, structured notes, or summaries. We similarly encode lightweight meeting or discussion context as structured notes when it is needed to preserve dependencies between consecutive tasks.

When a resource cannot be directly exposed or practically restored, we emulate access to it through a controlled mock API. The evaluated agent submits a query to the API in the form of a prompt, a structured request, or a code snippet. All such inputs are treated purely as text. A dedicated simulator agent then generates a task-relevant response, such as a statistic, summary, or simulated tool output, conditioned on the query and on known properties of the resource derived from the workflow records and human annotations. The responses are designed to be consistent with the role of the resource in the original workflow. This mechanism enables the evaluated agent to interact with otherwise unavailable resources without exposing private raw data or requiring the original resources to be restored and operated.

\paragraph{Reconstruction loop.}
We validate each reconstructed task through an execution loop. Given the generated instruction and reconstructed files or APIs, an agent executes the task in the prepared container environment. A verifier then compares the agent's result with the outcome recorded in the original worklog, judging whether they correspond to the same underlying task rather than requiring exact textual matching.

If the verifier determines that the execution corresponds to a different task, it produces revision feedback for refining the instruction, and the task is executed again. If the execution targets the correct task but fails during the process, we conduct human review to determine whether the reconstructed observations or environment are incomplete, and supplement them when necessary. Once the agent successfully performs the same task as reflected in the worklog, we retain the validated instruction, reconstructed observations, execution trajectory, and container state for benchmark construction. This loop ensures that each task is semantically faithful to the original workflow and executable in a controlled environment.

\subsection{Controlled State Progression and Trajectory Alignment}
\label{subsec:trajectory_alignment}

A challenge in constructing sequential agent benchmarks is that open-ended task execution can introduce execution drift. For many tasks, there may be multiple reasonable ways to complete the same user objective, and differences in intermediate decisions, file edits, or tool usage are not necessarily errors. However, if an agent freely executes the entire task stream from the beginning, these local differences can accumulate and alter future environments, making later tasks incomparable to the original workflow and difficult to evaluate consistently.

To prevent such drift, we align the validated trajectory obtained during task reconstruction to the real workflow outcome. Given the original pre-task and post-task workspace states, we rewrite the agent trajectory into a result-aligned trajectory. The rewritten trajectory preserves the high-level execution pattern of the agent, such as the sequence of observations, reasoning steps, tool calls, and file operations, but adjusts the concrete outputs so that applying the trajectory produces the same final workspace changes as recorded in the worklog. We then apply this aligned trajectory to update the container state, obtaining a canonical post-task environment for the next task.

This design gives every task a consistent pre-task history and execution environment, while still preserving realistic trajectory information from prior tasks. It also makes the benchmark more efficient: once canonical pre-task states are prepared, target tasks can be evaluated independently or in parallel. More importantly, by eliminating cumulative drift from earlier executions, the measured performance on each target task is more directly attributable to how the memory component uses fixed historical experience, rather than to uncontrolled variations in previous task execution.

\subsection{Metric Generation}

\label{subsec:evaluation_metrics}

\textbf{We prioritize evaluating whether memory improves downstream agent performance}. Therefore, we use two primary outcome metrics: Workspace Score, which measures whether the task is successfully completed, and Preference Score, which measures whether the agent preserves participant-specific working preferences. In addition, we introduce four diagnostic metrics, including Relevance, Continuity, Solvability and Hallucination Robustness, to better understand how memory influences agent behavior.

\paragraph{Preference Score.}
User preferences often reflect implicit requirements beyond task completion, such as collaboration habits and information organization styles. To capture preferences that persist across tasks, we construct participant-specific preference rubrics in two stages: human annotators first define root preferences, and an LLM analyzes historical message logs and document changes to derive additional cross-task rubrics. For each task, applicable rubrics are identified from the reference document changes, and an LLM judge evaluates whether the candidate trajectory's document writes satisfy these preferences. The weighted rubric scores are normalized to a 0--100 Preference Score. We additionally report an anonymized pairwise win rate between with-recall and without-recall trajectories under the same preference criteria.

\paragraph{Workspace Score.}
Workspace Score evaluates task completion based on the final Docker environment rather than the agent's response. Its reference is the validated post-task workspace obtained during task reconstruction. The evaluator derives task requirements from the reference workspace at three levels: minimum completion, reference-level quality, and above-reference improvements. Candidate code, documents, configurations, and artifacts are evaluated against these requirements, with additional execution checks applied when necessary. Scores are normalized to 0--100, where approximately 60 indicates minimum completion, 80 indicates reference-level quality, and higher scores require clear improvements in correctness, coverage, validation, usability, or robustness.

\paragraph{Memory Diagnostics.}
Workspace and Preference scores show whether memory improves the final result, but not how recalled history affects execution. We therefore examine both retrieval quality and its behavioral effects. Since real workflow traces do not specify which earlier tasks are useful, an LLM compares each evaluation task with all earlier tasks in the participant's workflow, identifies relevant predecessors, and extracts supporting excerpts from their agent traces. Together, they form the reference historical evidence. We organize our diagnostics around four questions:
\begin{itemize}
\item \textbf{Relevance: Does memory recall the history needed by the current task?}
We compute pairwise cosine similarities between recalled chunks and reference excerpts using an embedding model. Mean similarity averages the best match for each reference excerpt. Given a fixed threshold, recall and precision measure the matched proportions of reference excerpts and recalled chunks, respectively. In addition, an LLM judge assigns a 1--5 relevance score to the actual recall.

\item \textbf{Continuity: Does memory reduce repeated exploration?}
We use an LLM to classify every tool call as exploration (searching or reading), execution (modifying, running, or validating), or other. We compare their counts and proportions. In addition, an LLM assigns a 1--5 environment-familiarity score to the overall trajectory.

\item \textbf{Solvability: Can memory solve problems left unresolved without recall?}
An LLM reads the no-recall trajectory to identify execution problems and determine whether each is resolved by the end of the trajectory.
For each unresolved problem, another LLM judges whether the actual recall contains sufficient information to resolve or avoid it. We report the proportion of unresolved problems that are solvable using the recalled memory.

\item \textbf{Hallucination Robustness: Does memory cause new execution problems?}
To measure errors introduced by memory, an LLM first identifies execution problems from the with-recall trajectory. For each problem,
another LLM checks whether it is caused by misleading recall. We report the proportion of tasks containing at least one memory-induced problem. In addition, an LLM assigns a 1--5 hallucination-robustness
score to the overall trajectory.
\end{itemize}

\section{Empirical Findings}





\subsection{Data Statistics}
\label{sec:data-statistics}

\begin{figure}[t]
    \centering
    \includegraphics[width=\textwidth]
        {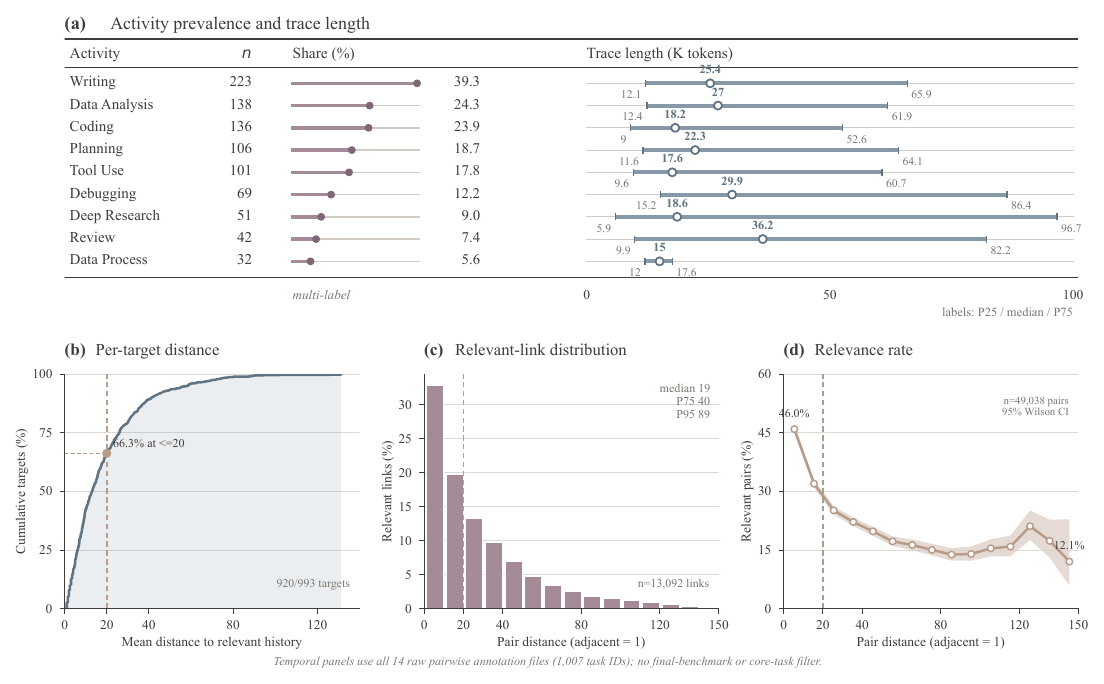}
    \caption{Task diversity and temporal relevance.
    (a) Multi-label activity prevalence and trace-length distributions.
    (b--d) Aggregated temporal-distance and relevance statistics.}
    \label{fig:task-diversity-and-temporal-relevance}
\end{figure}

\textbf{Benchmark scale.}
Following the data construction process described above, we constructed 1005 executable agent tasks from the long-term real-world workflows of 14 participants. Of these, 568 tasks were manually labeled as main tasks under the participants' primary project goals and served as core evaluation tasks; the remaining 437 tasks were background or temporary non-main tasks. Non-main tasks were not directly used as evaluation targets but were retained in the same participant's historical sequence to maintain noise and contextual dependencies in the real workflow. We used a GPT-5 tokenizer to calculate the size of the message log for each task. All 1005 message logs contained an average of 36.3K tokens, with the longest reaching 212.3K tokens.

\textbf{Task diversity.}
To characterize the diversity of task content, we manually labeled the agentic activities of 568 core tasks using multi-label annotation. As shown in Figure~\ref{fig:task-diversity-and-temporal-relevance}(a), these core tasks cover various real-world agent activities, including writing, data analysis, and coding. The trace size also varies depending on the task type; for example, debugging, review, writing, and deep research tasks typically contain longer contexts.

\textbf{Temporal relevance.}
For each core evaluation task, we used an LLM to identify relevant preceding tasks and extracted supporting evidence spans from their message logs to form gold historical evidence. Based on these annotations, we calculated the temporal distance between relevant preceding tasks and the current task. Of the 568 core tasks, 541 depended on at least one preceding task, resulting in 8084 relevant links. Among these, 75.6\% of the tasks had an average distance of no more than 20 subtasks between their relevant preceding tasks, demonstrating significant locality; however, 24.4\% still depended on earlier history with an average distance of more than 20 subtasks, indicating that the benchmark simultaneously includes short-term workflow continuity and long-term episodic dependency. Complete statistics can be found in Appendix~\ref{app:data-statistics}.



\subsection{How Different Memory Components Affect Agents}
\label{subsec:downstream_memory_evaluation}

This experiment evaluates whether memory components improve an agent's downstream performance in continuous real-world workflows. To evaluate different memory components under a controlled execution policy, we fix both the agent harness and the underlying model. We assess not only whether the agent produces an apparently complete response, but also whether the resulting artifacts are usable in the actual workspace and whether the agent preserves work practices established by the participant across earlier tasks.

\subsubsection{Evaluation Setting}

\paragraph{Execution Setting.}
 We use Codex\citep{openai2026codexcli} as the agent harness and GPT-5.5 with xhigh reasoning effort as the execution model. For each primary-workflow task, the agent starts from the canonical pre-task state produced by the trajectory-alignment procedure described in Section~\ref{subsec:trajectory_alignment} and executes independently in an isolated Docker environment. All conditions use the same task instruction, initial workspace, model, and tool permissions; they differ only in the recalled context provided to the agent.

\paragraph{Metric Setting.}
We evaluate downstream outcomes using the Workspace Score and Preference Score introduced in Section~\ref{subsec:evaluation_metrics}. We use GPT-5.5 for all grading calls and keep the grader model and prompts fixed across memory conditions. Workspace grading is performed through Codex with xhigh reasoning effort: the grader receives a task-specific evaluation bundle and read-only access to the candidate and reference Docker images, and is instructed to ground its score in inspected artifacts and executable checks. For Preference Score, each applicable participant-specific rubric is evaluated in an independent GPT-5.5 call. For the complementary pairwise evaluation, the with-recall and without-recall trajectories are randomly anonymized as candidates A and B, and the judge is required to select one candidate without ties. Detailed grading prompts are provided in Appendix~\ref{sec:app_prompt}. Table~\ref{tab:main_memory_results} reports the main results.

\paragraph{Diagnostic Setting.}
We use the Memory Diagnostics introduced in Section~\ref{subsec:evaluation_metrics} and keep the reference evidence, grader model, and prompts fixed across memory components. All LLM-based annotations and judgments use GPT-5.5. Reference historical evidence is constructed by comparing each evaluation task with all earlier tasks in the participant's workflow. Relevance, Continuity, and Hallucination Robustness are judged from the complete paired with-recall and without-recall trajectories. Tool-call classification, problem solvability, and memory-induced problems are evaluated through separate calls with fixed prompts. For the embedding-based relevance metrics, we encode recalled chunks and reference excerpts using \texttt{text-embedding-3-small} and apply a cosine-similarity threshold of 0.65. Detailed prompts are provided in Appendix~\ref{sec:app_prompt}. Table~\ref{tab:main_memory_results} reports the main results.

\subsubsection{Memory Turns Apparent Completion into Usable Work}

Table~\ref{tab:main_memory_results} shows that memory improves both the final workspace and alignment with participant-specific work practices. In particular, the Workspace Score increases from approximately 68 without memory to approximately 78 with the strongest memory component, moving performance from basic but unreliable completion toward reference-level usability. Preference follows the same overall trend.

Taken together, the Workspace and Preference results suggest that no-recall agents often achieve apparent completion without producing a usable, workflow-consistent outcome. To understand this pattern, we manually inspect the paired execution trajectories and their final workspaces. The resulting artifacts often appear plausible in isolation but fail to preserve the state and conventions established by prior work. Memory restores these latent task states and behavioral constraints, allowing the agent to continue an existing workflow rather than reconstruct a generic solution from the current instruction alone. Its primary value is therefore not to provide additional background knowledge, but to turn apparent completion into an output that is usable, verifiable, and consistent with the work that came before it.

\begin{table}[t]
    \centering
    \small
    \caption{Unified downstream and diagnostic results. We compare a \texttt{no-recall} baseline, in which no historical context is injected, with \texttt{with-recall} conditions augmented by different memory components. Column groups indicate the primary context representation used by each component. Diagnostic metrics are not defined for No Recall. Count and rate changes are computed as with-recall minus the shared without-recall trajectory; count changes are rounded to two decimal places and rate changes are reported in percentage points (pp).}
    \label{tab:main_memory_results}
    \resizebox{\textwidth}{!}{%
\begin{tabular}{lccccccc}
      \toprule
      & \multicolumn{1}{c}{\textit{Baseline}} & \multicolumn{4}{c}{\textit{Summary-Based Memory}} & \multicolumn{1}{c}{\textit{Task Summary}} & \multicolumn{1}{c}
      {\textit{ICL Experience}} \\
      \cmidrule(lr){2-2} \cmidrule(lr){3-6} \cmidrule(lr){7-7} \cmidrule(lr){8-8}
      \textbf{Metric} & \textbf{No Recall} & \textbf{mem0} & \textbf{memos} & \textbf{supermemory} & \textbf{memorybank} & \textbf{langmem} & \textbf{a-mem} \\
      \midrule
      \multicolumn{8}{l}{\textit{Downstream Performance}} \\
      Workspace Score $\uparrow$ & 68.08 & 72.48 & 70.01 & 70.57 & 73.24 & 75.29 & \textbf{78.20} \\
      Preference Score $\uparrow$ & 41.50 & 49.73 & 46.38 & 48.04 & 55.40 & 57.37 & \textbf{70.60} \\
      With-recall Win Rate $\uparrow$ & -- & 50.70\% & 53.30\% & 55.60\% & 65.08\% & 62.79\% & \textbf{72.70\%} \\
      \midrule
      \multicolumn{8}{l}{\textit{Relevance}} \\
      LLM Annotated $\uparrow$ & -- & 2.97 & \textbf{4.09} & 3.28 & 3.64 & 3.98 & 3.73 \\
      Mean Similarity $\uparrow$ & -- & 0.4825 & 0.5697 & 0.5095 & 0.5618 & 0.6153 & \textbf{0.7244} \\
      Recall $\uparrow$ & -- & 0.0757 & 0.2323 & 0.1076 & 0.2137 & 0.3687 & \textbf{0.6945} \\
      Precision $\uparrow$ & -- & 0.2190 & \textbf{0.3160} & 0.1842 & 0.2728 & 0.2209 & 0.3001 \\
      \midrule
      \multicolumn{8}{l}{\textit{Continuity}} \\
      LLM Annotated $\uparrow$ & -- & 3.13 & 3.20 & 3.15 & 3.44 & 3.93 & \textbf{4.00} \\
      Exploration $\Delta$ $\downarrow$ & -- & -0.79 & -1.20 & -0.77 & -1.19 & -2.57 & \textbf{-5.37} \\
      Execution $\Delta$ & -- & +0.74 & +1.41 & -0.17 & +2.19 & +1.66 & -0.09 \\
      Exploration Rate $\Delta$ $\downarrow$ & -- & -2.22 pp & -4.27 pp & -0.69 pp & -5.59 pp & -6.33 pp & \textbf{-7.06 pp} \\
      Execution Rate $\Delta$ $\uparrow$ & -- & +2.38 pp & +3.93 pp & +0.40 pp & +5.37 pp & +6.22 pp & \textbf{+6.63 pp} \\
      \midrule
      \multicolumn{8}{l}{\textit{Hallucination Robustness}} \\
      LLM Annotated $\uparrow$ & -- & \textbf{4.90} & 4.85 & 4.87 & 4.83 & 4.76 & 4.66 \\
      Memory-induced Task Rate $\downarrow$ & -- & \textbf{0.35\%} & \textbf{0.35\%} & 0.70\% & 1.23\% & 5.11\% & 7.39\% \\
      \midrule
      \multicolumn{8}{l}{\textit{Solvability}} \\
      Problem Solvability Rate $\uparrow$ & -- & 1.45\% & 1.45\% & 0.96\% & 1.45\% & 6.59\% & \textbf{7.07\%} \\
      \bottomrule
  \end{tabular}}
  \end{table}

\subsubsection{In-Context Experience Is More Effective than Summarization}

Table~\ref{tab:main_memory_results} highlights the distinction between summary-oriented memory and in-context experience. Summary-based systems perform well on semantic relevance and precision but remain weaker in evidence coverage, solvability, and downstream performance. In contrast, \texttt{a-mem}, a representative of ICL-style memory, provides broader historical coverage, stronger continuity, and a larger reduction in exploration. The structured task-level summaries produced by \texttt{langmem} fall between these two regimes. Overall, the closer recalled history is to a concrete working example, the less context the agent must reconstruct before acting.

The Solvability results provide a direct explanation for this pattern: ICL-style memory and structured task summaries cover a substantially larger share of execution problems for which historical evidence is useful than the lighter-weight summary baselines. Summarization primarily preserves task conclusions and state, whereas in-context experience additionally retains the paths, constraints, existing artifacts, and procedures needed to act on that history. This actionability comes with a trade-off, as richer recall is also associated with more memory-induced problems. The results therefore favor curated in-context experience over pure summarization, rather than indiscriminately retaining more history. Because the components also differ in retrieval strategy and context length, we interpret this comparison as an empirical association and leave a controlled study under a fixed token budget to future work.

\subsection{Different Base Models Use Memory Differently}
\label{subsec:base_model_evaluation}

Having compared memory components under a fixed execution model, we next examine how memory interacts with different base models. We ask whether models benefit equally from recall, how recall changes their execution behavior, and how reliably they use the information it provides.

\subsubsection{Evaluation Setting}

We follow the execution setting in Section~\ref{subsec:downstream_memory_evaluation}, using Codex as the agent harness and the same task instructions, canonical pre-task workspaces, and tool permissions. We fix the memory component to \texttt{mem0} and vary only the execution model, evaluating DeepSeek-V4-Pro, GPT-5.5, GLM-5.1, Kimi-K2.6, and Qwen3.7-Max under both no-recall and with-recall conditions on all 568 core tasks.

We use the Workspace, Preference, Continuity, and Hallucination Robustness metrics described in Sections~\ref{subsec:downstream_memory_evaluation} and~\ref{subsec:evaluation_metrics}. All grading, annotation, and judgment calls use GPT-5.5 with fixed prompts. We additionally examine which unresolved problems could have been solved using recall. GPT-5.5 identifies unresolved problems in each with-recall trajectory, and a separate GPT-5.5 call checks whether the recall contains enough information to solve or avoid each problem. Recall-Solvable Problems is the percentage of unresolved problems that could have been solved using the recall; lower is better. Table~\ref{tab:base_model_results} reports the results.

\begin{table}[t]
    \centering
    \small
    \caption{Downstream performance and memory-use diagnostics across base models with \texttt{mem0}. All changes are computed as with-recall minus no-recall; count changes are rounded to two decimal places and rate changes are reported in percentage points (pp).}
    \label{tab:base_model_results}
    \resizebox{\textwidth}{!}{%
    \begin{tabular}{lccccc}
        \toprule
        \textbf{Metric} & \textbf{DeepSeek-V4-Pro} & \textbf{GPT-5.5} & \textbf{GLM-5.1} & \textbf{Kimi-K2.6} & \textbf{Qwen3.7-Max} \\
        \midrule
        \multicolumn{6}{l}{\textit{Downstream Performance}} \\
        Workspace Score, No Recall $\uparrow$ & 61.76 & 67.53 & \textbf{71.91} & 66.18 & 62.93 \\
        Workspace Score, With Recall $\uparrow$ & 67.36 & 72.48 & \textbf{74.10} & 69.16 & 65.99 \\
        Workspace Score $\Delta$ $\uparrow$ & \textbf{+5.61} & +4.95 & +2.19 & +2.99 & +3.06 \\
        Preference Score, No Recall $\uparrow$ & 32.48 & 42.07 & \textbf{46.09} & 40.20 & 35.83 \\
        Preference Score, With Recall $\uparrow$ & 42.09 & 49.73 & \textbf{51.91} & 48.57 & 41.39 \\
        Preference Score $\Delta$ $\uparrow$ & \textbf{+9.61} & +7.66 & +5.83 & +8.37 & +5.55 \\
        With-recall Win Rate $\uparrow$ & \textbf{61.09\%} & 50.70\% & 55.46\% & 57.39\% & 51.94\% \\
        \midrule
        \multicolumn{6}{l}{\textit{Continuity}} \\
        LLM Annotated $\uparrow$ & 3.19 & 3.13 & \textbf{3.33} & 3.29 & 3.22 \\
        Exploration $\Delta$ $\downarrow$ & +0.26 & -0.32 & -2.49 & \textbf{-3.01} & +0.37 \\
        Execution $\Delta$ $\uparrow$ & \textbf{+2.28} & +0.27 & -1.07 & -0.76 & +0.56 \\
        Exploration Rate $\Delta$ $\downarrow$ & \textbf{-3.40 pp} & -0.79 pp & -0.68 pp & -1.20 pp & -1.06 pp \\
        Execution Rate $\Delta$ $\uparrow$ & \textbf{+3.26 pp} & +0.94 pp & -0.05 pp & +0.49 pp & +1.12 pp \\
        \midrule
        \multicolumn{6}{l}{\textit{Hallucination Robustness}} \\
        LLM Annotated $\uparrow$ & 4.32 & \textbf{4.90} & 4.35 & 4.41 & 4.31 \\
        Memory-induced Task Rate $\downarrow$ & 0.53\% & \textbf{0.00\%} & 2.11\% & 0.88\% & 1.23\% \\
        \midrule
        \multicolumn{6}{l}{\textit{Memory Utilization}} \\
        Recall-Solvable Problems $\downarrow$ & 1.89\% & \textbf{1.52\%} & 2.27\% & 3.03\% & 2.23\% \\
        \bottomrule
    \end{tabular}}
\end{table}

\subsubsection{Memory Improves Every Model, but by Different Amounts}

Table~\ref{tab:base_model_results} shows that \texttt{mem0} improves both Workspace and Preference scores for every model. However, the model ranking is already largely established without recall and remains similar after memory is added. GLM achieves the highest Workspace score in both conditions, whereas DeepSeek obtains the largest Workspace improvement. The same pattern appears in Preference: every model benefits from recall, but the magnitude of the gain varies substantially. Memory therefore provides useful historical state without removing differences in the models' underlying ability to complete tasks.

\subsubsection{Models with Larger Gains Explore Less and Execute More}

The behavioral results distinguish reducing tool use from using memory productively. Kimi and GLM show the largest reductions in the absolute number of exploration calls, but their execution calls also decrease and their Workspace gains are relatively small. DeepSeek instead shifts its tool-call distribution: its Exploration Rate decreases by 3.40 percentage points while its Execution Rate increases by 3.26 points, accompanied by the largest Workspace gain of 5.61. GPT shows the same direction with a smaller behavioral shift and the second-largest Workspace gain. Memory is most useful when it reallocates effort from reconstructing workspace state to modifying, running, and validating task outputs, rather than merely shortening the trajectory.

\subsubsection{Models with Larger Gains Use Recall Better}

GPT and DeepSeek have the lowest rates of Recall-Solvable Problems, at 1.52\% and 1.89\%, and also obtain the two largest Workspace gains. Qwen and GLM fall in the middle, at 2.23\% and 2.27\%, while Kimi has the highest rate at 3.03\%. This pattern suggests that one source of variation in memory gains is whether a model turns recalled information into effective action: recall may already contain a useful path or constraint, yet the corresponding problem can remain unresolved. Because the memory component is fixed to \texttt{mem0}, this metric focuses on differences in how base models act on the provided recall rather than differences between memory systems.

\subsubsection{Some Models Are More Easily Misled by Memory}

The model also affects the risk introduced by memory. Although every condition uses \texttt{mem0}, Memory-induced Task Rate ranges from 0.00\% for GPT to 2.11\% for GLM. These cases include all execution problems caused or worsened by misleading recall, regardless of whether the agent later resolves them. Memory robustness is therefore determined jointly by what the memory system retrieves and how the base model interprets, verifies, and applies that information.

\section{Conclusion}

We introduced \benchmarkname, a longitudinal benchmark that transforms privacy-preserved real-world workflows into 1,005 executable tasks with controlled environments and downstream-oriented evaluation, including 568 core tasks. Experiments across six memory components and five base models show that recall improves workspace quality and preference alignment in the tested settings, but that its value depends on whether the retrieved experience is actionable and whether the base model can apply it reliably. By jointly measuring task outcomes and the relevance, continuity, solvability, and robustness of recalled history, \benchmarkname{} provides a reproducible basis for developing memory systems that support sustained workflow execution.

\section*{Limitations and Future Work}
We designed MemoryBench with the goal of making memory evaluation as comprehensive as possible while remaining faithful to real-world agent workflows. Despite our efforts, the current benchmark still has several limitations. Below, we summarize the limitations that we have identified and outline our immediate priorities for improving the benchmark.

\subsection*{Rubric Calibration and Human Validation}

Calibration of the task-specific evaluation rubrics is still ongoing. Rather than relying primarily on fixed statistical metrics, which are generally stable but limited in their ability to capture open-ended task quality, we use rubric-based grading as the main evaluation protocol. This choice reflects the fact that success in many real-world tasks is inherently subjective and multidimensional. However, it also requires careful calibration to ensure that model-based grading reliably reflects human judgment.

As an initial validation, we have collected coarse-grained human judgments for a sampled subset of tasks and compared them with the corresponding model-based evaluations. The two exhibit strong preliminary agreement. Nevertheless, this analysis does not yet provide exhaustive validation at the level of individual tasks and rubric criteria. We plan to expand the human evaluation to systematically calibrate each task-specific rubric and investigate disagreement patterns between human and model graders.

\subsection*{Evaluation Cost and Scalable Proxies}

We have made substantial efforts to reduce the cost of evaluating memory in realistic settings---from a process that might otherwise require deploying a product and conducting online experiments to one that can be executed in a controlled sandbox. Nevertheless, the resulting evaluation remains relatively expensive: under the current setup, a complete benchmark run costs approximately \$200 per model--harness configuration.

We plan to reduce this cost by constructing representative task subsets and streamlining selected task environments while preserving their objectives and memory dependencies. Our goal is to provide a lower-cost proxy evaluation that supports rapid experimentation, while retaining the full benchmark for comprehensive evaluation.

\subsection*{Model and Agent-Harness Coverage}

The current results do not yet cover several recently released models or a sufficiently broad range of agent harnesses. Consequently, they should not be interpreted as an exhaustive comparison of existing memory-enabled agent systems. We plan to continuously expand the evaluation to include newer models, additional agent harnesses, and a broader range of memory implementations.

\subsection*{Broader Domains and Richer Observations}

MemoryBench is constructed primarily from real-world workflow data, which is difficult to access and often contains sensitive information. As a result, the current benchmark covers only a limited set of participants, domains, and observation types. This limitation may constrain the extent to which the results generalize to other professional workflows and interaction settings.

We hope to collaborate with additional open-source projects and data contributors who are willing to share appropriately consented and anonymized process-level data. Such contributions would enable us to extend MemoryBench to broader domains, richer forms of observation, and more diverse patterns of long-term dependency.

\clearpage
\bibliographystyle{unsrtnat}
\bibliography{references}

\appendix
\appendixpage
\numberwithin{figure}{section}
\numberwithin{table}{section}
\numberwithin{equation}{section}
\section{Data Statistics}
\label{app:data-statistics}
We report the token distribution of converted message logs, anonymized participant-level history lengths, multi-label task type annotations, token statistics by task type, and the temporal locality of relevant historical tasks. These statistics complement Section~\ref{sec:data-statistics} by giving more detailed evidence about the scale, diversity, and long-horizon dependency structure of the benchmark.

\begin{table}[h]
  \centering
  \caption{Token statistics of message logs.}
  \label{tab:message-log-token-stats}
  \begin{tabular}{C{0.9cm}C{1.0cm}C{1.0cm}C{1.3cm}C{1.0cm}C{0.9cm}C{1.1cm}}
  \toprule
  n & Mean & P25 & Median & P75 & Min & Max \\
  \midrule
  1005 & 36.3K & 8.4K & 21.2K & 56.1K & 641 & 212.3K \\
  \bottomrule
  \end{tabular}
\end{table}

\begin{table}[h]
    \centering
    \caption{Anonymized participant-level pre-task history token statistics. Participants are grouped by the number of core evaluation tasks. For each core target task, history includes all previous sessions from the same participant, including
    both core and non-core tasks.}
    \label{tab:participant-history-token-stats}
    \resizebox{\textwidth}{!}{%
    \begin{tabular}{lrrrrrrrrr}
    \toprule
    Group & Ppl. & All & Core & Non-core & Mean & P25 & Median & P75 & Max \\
    \midrule
    0--9 core tasks & 1 & 4 & 3 & 1 & 4.3K & 1.9K & 3.9K & 6.5K & 9.0K \\
    10--19 core tasks & 2 & 38 & 32 & 6 & 82.1K & 18.3K & 48.8K & 109.5K & 351.2K \\
    20--39 core tasks & 4 & 226 & 104 & 122 & 927.3K & 183.6K & 685.1K & 1.54M & 3.13M \\
    40--59 core tasks & 3 & 251 & 155 & 96 & 917.5K & 141.7K & 576.3K & 1.18M & 4.01M \\
    60--79 core tasks & 4 & 486 & 274 & 212 & 1.59M & 255.4K & 962.6K & 2.25M & 8.43M \\
    \midrule
    Overall & 14 & 1005 & 568 & 437 & 1.19M & 153.6K & 654.1K & 1.77M & 8.43M \\
    \bottomrule
    \end{tabular}}
\end{table}

\begin{table}[h]
  \centering
  \caption{Multi-label task tag distribution over 568 core tasks. }
  \label{tab:task-tag-distribution}
  \begin{tabular}{L{3.0cm}C{2.0cm}C{1.5cm}}
  \toprule
  Tag & n & \% of 568 \\
  \midrule
  Writing & 223 & 39.26 \\
  Data Analysis & 138 & 24.30 \\
  Coding & 136 & 23.94 \\
  Planning & 106 & 18.66 \\
  Tool Use & 101 & 17.78 \\
  Debugging & 69 & 12.15 \\
  Deep Research & 51 & 8.98 \\
  Review & 42 & 7.39 \\
  Data Process & 32 & 5.63 \\
  \bottomrule
  \end{tabular}
\end{table}

\begin{table}[h]
  \centering
  \caption{Message log token statistics by task type. }
  \label{tab:task-type-token-stats}
  \begin{tabular}{L{2.6cm}C{1.0cm}C{1.0cm}C{1.0cm}C{1.3cm}C{1.0cm}C{1.1cm}}
  \toprule
  Tag & Tasks & Mean & P25 & Median & P75 & Max \\
  \midrule
  Writing & 223 & 46.9K & 12.1K & 25.4K & 65.9K & 212.3K \\
  Data Analysis & 138 & 44.6K & 12.4K & 27.0K & 61.9K & 184.1K \\
  Coding & 136 & 36.1K & 9.0K & 18.2K & 52.6K & 188.3K \\
  Planning & 106 & 42.2K & 11.6K & 22.3K & 64.1K & 188.3K \\
  Tool Use & 101 & 42.7K & 9.6K & 17.6K & 60.7K & 188.9K \\
  Debugging & 69 & 51.5K & 15.2K & 29.9K & 86.4K & 183.0K \\
  Deep Research & 51 & 46.6K & 5.9K & 18.6K & 96.7K & 185.8K \\
  Review & 42 & 49.9K & 9.9K & 36.2K & 82.2K & 167.4K \\
  Data Process & 32 & 21.2K & 12.0K & 15.0K & 17.6K & 112.5K \\
  \bottomrule
  \end{tabular}
\end{table}

\begin{table}[h]
  \centering
  \caption{Temporal locality of relevant previous tasks. Among 568 core tasks, 541 have at least one relevant previous task, with 8084 relevant links in total.}
  \label{tab:relevance-locality}
  \begin{tabular}{L{3.1cm}C{1.5cm}C{2.0cm}}
  \toprule
  Mean subtask-id distance & Subtasks & \% among nonzero \\
  \midrule
  0--10 & 261 & 48.24 \\
  10--20 & 148 & 27.36 \\
  20--50 & 114 & 21.07 \\
  50+ & 18 & 3.33 \\
  \bottomrule
  \end{tabular}
\end{table}

\FloatBarrier
\section{Prompts and Annotation Protocol}
\label{sec:app_prompt}

This appendix reports the prompt templates and human annotation protocol used in our data construction pipeline. 

\subsection{Task Segmentation}
\label{sec:app_prompt_segmentation}

We first apply a rule-based temporal pre-merge to nearby document-editing events. An LLM assistant then decides whether each candidate temporal chunk should be merged into the current task or split into a new task. The prompt emphasizes fine-grained task boundaries and treats over-merging as more harmful than mild over-splitting.

\begin{promptbox}{Instruction Generation Prompt}
You are given a segmented worklog entry together with its associated
You are merging consecutive temporal worklog chunks into larger tasks.

You are given:
1) the current merged worklog group,
2) one candidate next temporal chunk,
3) a short lookahead of later temporal chunks.

Important setup:
- Before you see the payload, a rule-based pre-merge has already grouped very close-in-time edits into temporal chunks.
- These temporal chunks are only there to reduce merge pressure on the model.
- The final saved task will still keep the finest-grained original event ids.
- So your job is only to decide whether the candidate chunk should join the current task, not to reason about UI granularity.

Goal:
- Decide whether the candidate next temporal chunk should be merged into the current task.
- Prefer a fine-grained segmentation.
- Be actively willing to split tasks.
- Unless there is strong evidence that the candidate chunk is still part of the same concrete task, lean toward SPLIT rather than over-merging.
- When uncertain, prefer SPLIT over MERGE.
- In this task, over-merging is worse than mild over-splitting.
- Be comfortable splitting aggressively when there is any plausible boundary.
- Do not treat MERGE as the default. Treat SPLIT as the safer default unless the evidence for MERGE is genuinely strong.
- Do not keep extending the same merged group by default. Re-check the boundary fresh at every step.
- Especially when the current merged group is already long or contains many accumulated edits, be more willing to SPLIT rather than continue merging.
- A task should usually correspond to one coherent goal / artifact / investigation / pipeline stage, not one tiny worklog edit.

Payload field meanings:
- temporal\_premerge.median\_gap\_seconds / merge\_gap\_threshold\_seconds: the rule-based timing statistics used to build temporal chunks. These are only context; they are not direct merge commands.

- group\_chunk\_count: how many temporal chunks are already inside the current merged group before considering the candidate.

- group\_event\_count: how many raw worklog events are already inside the current merged group before considering the candidate.

- large\_group\_split\_hint: a stronger reminder string that appears when the current merged group is already large (>15 events). Treat it as a prompt to actively inspect the current boundary and look for a reasonable place to split.

- group\_markdown\_view.group\_start\_markdown: the full markdown snapshot from the beginning of the current merged group, before the first event in the group happened.

- group\_markdown\_view.current\_markdown: the full markdown snapshot of the current merged group after all chunks already in the group have been applied. This is the main current state of the merged group.

- group\_markdown\_view.candidate\_after\_markdown: the full markdown snapshot you would get if the candidate chunk were merged into the group. Compare this against current\_markdown to judge whether the candidate still looks like the same task.

- group\_markdown\_view.large\_markdown\_omitted / omission\_note: when the full markdown views would be too large, some or all of them may be intentionally omitted.

- group\_markdown\_view.before\_omitted: when true, group\_start\_markdown was omitted.

- group\_markdown\_view.candidate\_after\_omitted: when true, candidate\_after\_markdown was omitted.

- group\_markdown\_view.local\_only\_mode: when true, all full markdown views were omitted, including the before view. In that case, rely only on local chunk markdown and change summaries.

- group\_chunks: concise summaries of the temporal chunks already inside the current merged group.

- candidate\_chunk.markdown\_view.before\_local\_markdown: local markdown near the candidate chunk's focus path before that chunk happens.

- candidate\_chunk.markdown\_view.after\_local\_markdown: local markdown near the candidate chunk's focus path after that chunk happens.

- candidate\_chunk.change\_view: auxiliary structured metadata about the candidate chunk, including the raw events inside it.

- lookahead\_chunks[*].markdown\_view.before\_local\_markdown / after\_local\_markdown: local markdown around the next few future temporal chunks. These are only short future hints, not the main evidence.

Important:

- Do not force everything under one broad heading to stay together. If the work under the same heading clearly branches into different concrete tasks, you may split it.

- It is acceptable to produce more tasks when that makes each task map more cleanly to one concrete thing a person or agent would actually do.

- If merging would create one broad mixed bundle that really contains multiple plausible standalone tasks, choose SPLIT.

- A large existing merged group is itself a warning sign against further merging. Do not let momentum alone keep unrelated or weakly related edits inside the same task.

- If `group\_event\_count` is already greater than 15, explicitly ask whether the group has already become too broad and whether a split is now warranted.

- If `large\_group\_split\_hint` is present, do not passively note it; actively inspect the current boundary and look for a reasonable split point.

- Prefer group\_markdown\_view.current\_markdown and group\_markdown\_view.candidate\_after\_markdown plus the candidate\_chunk local BEFORE/AFTER markdown as the primary evidence.

- Treat focus paths and change summaries only as auxiliary hints.

- If a heading, bullet, or label clearly introduces content that continues immediately after it, keep that continuation together with the introduced content.

- In particular, patterns like "Prompt design: ", "Result: ", "Problem: ", "Development process: ", or similar label-plus-colon structures usually belong with the text or list items that immediately follow, so do not split them apart unless a clearly different task begins later.

- Use lookahead carefully when a date heading is still being formed.

- If a genuinely new date bucket appears, split immediately at that point.

- THIS RULE IS IMPORTANT: a genuinely new date bucket starts a new task immediately.

- If you say in the reason that the candidate introduces a genuinely new date section / new day bucket, then the final decision must be SPLIT, not MERGE.

- This date rule has higher priority than "the current group is still small", "the work is still related", "lookahead continues the same thread", or any other merge-leaning heuristic.

- So even if the current merged group is tiny, even if the broader topic is the same, and even if the lookahead keeps filling the new date section, a genuinely new date bucket must still SPLIT.

- If the markdown delta is subtle or nearly invisible, use the change-view / focus-view to understand what was edited.

- The payload may explicitly tell you whether the normalized BEFORE/AFTER markdown is identical; if so, that is a very strong signal to merge, because the edit likely did not change substantive content.

- If the markdown shows no substantive content update, prefer merging by default.

- A temporal chunk may contain multiple tiny edits that happened very close in time. Do not split just because the chunk itself contains multiple raw events. Judge whether the candidate chunk, as a whole, still belongs to the same concrete task.

- Pure formatting changes, structural cleanup, heading reshaping, or other non-content edits usually should not start a new task on their own.

- If the previous edit introduced or reshaped a large block of content and the candidate mainly edits a smaller piece inside that same newly introduced block, that usually should MERGE.

- In other words: "big block added, then a local refinement inside that block" is normally one continuing task, not a new one.

- A small current merged group is only a weak secondary heuristic. It must not outweigh clear evidence for SPLIT.

- Early in a task, sparse context is not by itself a reason to merge. If the candidate already looks like a separate plausible task, choose SPLIT.

- Conversely, when the current merged group is already large, you should require strong positive evidence to keep merging. If that strong evidence is missing, choose SPLIT.

- Once the merged group becomes large, the burden of proof flips: do not ask "can I keep merging?" Ask "is there strong evidence that this still must stay together?" If not, output SPLIT.

- For evaluation / benchmarking / model-comparison work, a model switch often suggests a new task, because the target under evaluation may have changed even if the surrounding section title remains similar.

- Treat this as a soft hint rather than a hard rule: use the full markdown context to decide whether the work is still one coherent task or has shifted to a new evaluation target.

- If the edit mainly creates a new date section such as "8.27", "9.14", or similar day markers, and that is a genuinely new day subsection rather than a revision of an old heading, that new date section itself should be treated as the new task boundary.

- Do not write a reason like "this is a genuinely new date section, but still merge because it is part of the same weekly thread". That is incorrect for this task.

- But once you are already inside one date bucket, that day can still contain multiple tasks.

- If one day clearly contains multiple different goals, deliverables, investigations, or pipeline stages, you should still split within that day at the appropriate boundary.

- So the same date can contain multiple tasks, but introducing a genuinely new date bucket is itself already a split signal.

- When the candidate touches dates or day buckets, structure your reason in this order:

- First, explicitly state whether the candidate introduces a genuinely new date section or merely edits an existing one.

- Second, explicitly state whether the lookahead shows continued content under that same new date section.

- Third, state the merge / split conclusion.

- Keep this reasoning concise, but do include these date checks in the reason whenever dates are involved.

- End the reason with exactly one final sentence in this exact format:

- `FINAL DECISION: I will MERGE.`

- or `FINAL DECISION: I will SPLIT.`

- Do not use any other final wording, and do not include both.

- Return STRICT JSON only.

Date Rules:

- If the edit is mainly renaming an existing heading into a date label, correcting an existing date label, or otherwise revising a date marker that already belongs to the current thread, that usually should MERGE with the current task.

- If the candidate chunk creates a genuinely new date subsection / new day bucket, it MUST start a new task immediately.

- A transition like "8.19" -> "8.20" is therefore itself a split boundary when 8.20 is a genuinely new date section.

- Do not override it because the current group is small.

- Do not override it because the candidate looks like a continuation of the previous day's work.

- Do not override it because the lookahead keeps adding content under the new date.

- If it is a genuinely new date bucket, output SPLIT.

- When judging this, compare current\_markdown with candidate\_after\_markdown: if the candidate clearly opens a new day container rather than merely revising the existing one, output SPLIT.

\end{promptbox}

\subsection{Human Annotation Protocol}
\label{sec:app_human_annotation}

After LLM-assisted segmentation, annotators reviewed the candidate tasks through an annotation interface. This step did not use an LLM prompt. Annotators followed a simplified standard operating procedure:
\begin{enumerate}
    \item Inspect each candidate task together with its document snapshots, local markdown diff, temporal context, and neighboring tasks.
    \item Verify whether the candidate should remain separate or be merged with adjacent tasks. The target unit is a coherent and relatively atomic work objective that could naturally be delegated to an agent.
    \item Mark whether the task belongs to the participant's main workflow.
    \item Write a concise task description that captures the work objective, boundary, and expected outcome.
    \item Write a metric description that identifies the evaluation focus for the task.
    \item Assign applicable agentic tags, such as coding, planning, debugging, data analysis, writing, tool use, review, or deep research.
    \item Select applicable participant-level preference metrics.
    \item Mark whether additional files or intermediate resources are needed. If so, describe the required resource traits, format, granularity, and content pattern.
\end{enumerate}

\begin{promptbox}{Instruction Generation Prompt}

\{

  "annotation\_status": "",

  "task\_description": "",

  "metric\_description": "",

  "selected\_preference\_metric\_ids": [],

  "agentic\_tags": [],

  "file\_supplement\_needed": "",

  "file\_supplement\_traits": "",

  "updated\_at": ""
  
\}
\end{promptbox}

\subsection{Instruction Generation and Task Reconstruction}
\label{sec:app_prompt_instruction_generation}

Given a segmented worklog entry, we infer the underlying work objective and then convert it into an executable agent task. The first prompt asks the model to infer what work likely produced the observed document change. The second prompt converts that inferred work into a natural agent instruction with resource dependencies, success criteria, expected outputs, and caveats.

\begin{promptbox}{Instruction Generation Prompt}
You will receive the following information:

- The main task currently being handled by the user

- Task comments for the current subtask

- A partial snippet of `before\_markdown` / `after\_markdown`

- A brief summary of the markdown diff

- A summary of the historical work and most recent version related to this document prior to the current task. This summary comes from the incremental document state maintained in subtask order.

- The markdown diff itself, if necessary.

Your task is not to write agent instructions, but to infer: what work was actually done this time.

Requirements:

- You need to infer "what the executor most likely did to produce this document change," not just restate the results already written in the document.

- You need to combine the main task, `task\_description`, and document differences to make this judgment.

- If historical context of "what was done before the current task" is provided, treat it as an important clue: the current subtask is often a continuation, supplementary verification, summary, or status confirmation based on this existing work.

- If the `task\_description` is "summary/organization/induction", but the document differences show new additions such as research results, example summaries, rule summaries, or status confirmations, then you should try to revert to the original description of "first conduct research/verify/design/screen, then form a summary or conclusion," rather than simply "organize existing content."

- `summary` should be a concise summary of the actual work done.

- `primary\_goal` should describe the core objective of the work itself, not the document update objective.

- `likely\_actions` should list 2-5 high-level actions, describing what is typically done to complete this task.

- `result\_signal` should summarize the resulting signals from this document change.

- `notes` should only contain necessary conservative explanations or points of uncertainty.

- If the evidence is insufficient, set `status` to `insufficient\_evidence`, but still try to provide conservative inferences.

Returns STRICT JSON, returning only one JSON object.

\end{promptbox}

\begin{promptbox}{Instruction Generation Prompt}
You will receive:

- A pre-inferred `inferred\_work`

- A list of optional middleware interfaces

Your job is to transform the `inferred\_work` into a fully executable agent task.

Requirements:

- The output `agent\_task.instruction` should be natural, specific, and executable.

- The instruction should read like a person handing over a task to another agent, not like a data summary or debriefing report.

- The instruction should be a high-level task handover, not written like an SOP, troubleshooting manual, or step-by-step guide.

- If this is an inspection/analysis/verification task, you can point out 2-4 aspects to focus on, but don't specify overly detailed execution orders, interface call rounds, or question templates.

- If the input provides interfaces, the instruction should naturally explain "what information you can obtain using these interfaces," but don't create new interfaces.

- Don't directly copy the final before/after text; extract the underlying task objectives and state changes.

- Avoid using specific target document wording in the instruction; do not write final, text-level statements like "change the status from A to B." Only abstract expressions like "determine whether a status upgrade/tightening/supplementing stronger conclusions are supported."

- Key point: `agent\_task` describes what the task itself needs to accomplish, not how the document will ultimately be written. Document modifications are merely traces of the completed task.

- For data inspection tasks, the real goal should be "determining whether the data is usable, its quality, and what the limitations are," not "determining whether it is sufficient to support document updates."

- If evidence is conflicting or insufficient, be conservative, but avoid vague statements.

- Only restore `agent\_task`; do not output `writeback\_task` or `evidence\_summary`. How the document is written back is not part of the current fixed process.

- `inferred\_work.primary\_goal` describes the actual task objective; prioritize it.

- `inferred\_work.likely\_actions` is for high-level reference only; do not mechanically copy it as a checklist.

Good style example of instruction:

"Examine existing GPT distillation data to determine its current usability and whether there are any constraints that require retention. You need to focus on overall usability, consistency of output quality, whether there are obvious problem patterns, and whether the existing evidence is sufficient to support a more explicit quality judgment. You can use the query\_xxx interface to obtain the overall conclusion, main problems, and coarse quantification basis. Finally, give a clear but conservative usability conclusion and supplement it with a minimal evidence summary."

Returns STRICT JSON, returning only a single JSON object.
\end{promptbox}

\subsection{Meeting Discussion Rewrite}
\label{sec:app_prompt_meeting_rewrite}

Some static resources represent informal meetings or discussion records. We rewrite these records into a consistent dialogue style while preserving task-relevant facts.

\begin{promptbox}{Meeting rewrite prompt}
You are editing resource JSON files inside one Membench resource `trace` directory.

Current working directory:
\{trace\_dir\}

Candidate JSON files:
\{candidate\_paths\}

Primary meeting files detected by code:
\{primary\_meeting\_paths\}

Primary meeting hint note:
\{primary\_meeting\_hint\_note\}

Task:
Some meeting files are detected by code: a JSON object with a top-level `messages` array whose items have this shape:
`\{\{"speaker": "...", "content": "..."\}\}`
Those files are definitely meeting/discussion records.

The code-detected list is only a hint. Even if it is empty, you must still inspect all candidate JSON files in this same resource directory to see whether any meeting files were missed. A missed meeting file must still have explicit speaker/content style dialogue, for example a list of objects with speaker-like and content-like fields. If a file has no speaker/person field paired with spoken content/text, it is not a meeting file.

If you do not find any meeting file, do not modify any file. End with no changes. This is mandatory.

You may freely rewrite the meeting dialogue structure inside the message list: split one long message into several shorter turns, merge adjacent turns when natural, reorder nearby turns if it makes the conversation flow better, and adjust speaker turn distribution. The goal is to make the dialogue read like a real meeting conversation instead of stiff reconstructed notes.

Output shape requirement for every meeting file you rewrite:
The final JSON file must contain only one top-level key: `messages`. Delete every other top-level key, including fields such as `record\_type`, `topic`, `participants`, `student`, `time\_window`, `time\_scope`, `source\_basis`, `level\_notes`, `constraints`, `notes`, `summary`, `metadata`, and any similar non-message fields. Do not preserve the original top-level schema. Before deleting those fields, inspect them: if they contain meaningful facts, constraints, dates, participants, source notes, task decisions, TODOs, blockers, or other useful information, fold that information naturally into the dialogue turns in `messages`. If a field is only administrative bookkeeping and contains no useful information for the later agent, discard it. After rewriting, the file should look like `\{\{"messages": [\{\{"speaker": "...", "content": "..."\}\}, ...]\}\}` and nothing else.

This is especially important for structured context files where useful information may live outside `messages`, for example keys like `strategy\_chain`, `dedup\_policy`, `script\_anchors`, `observed\_dedup\_statistics`, `prompt\_anchors`, `rules`, `requirements`, `statistics`, `examples`, or other nested objects and lists. You must not leave those as separate keys. Read them, decide which facts matter for a later agent, and turn only the useful task-relevant facts into natural meeting dialogue. If a field or nested value is unimportant, redundant, purely administrative, or not needed for understanding the task, it is better to discard it than to force it awkwardly into the conversation. Do not paste JSON-like bullet lists into a single message, and do not write stiff lines like "metadata says..." or "the strategy\_chain is...". Instead, make it sound like real people are talking through the information: one person asks or confirms, another person explains, they clarify important numbers, file names, thresholds, model names, and constraints in short conversational turns. The final file must still contain only `messages`.

The one thing you must not change is the useful task meaning. Preserve task-relevant decisions, facts, constraints, task details, names, dates, numbers, model names, file paths, URLs, examples, conclusions, TODOs, blockers, and assignments when they matter for understanding or executing the task. Do not add new information. You may discard unimportant administrative details or redundant metadata that a later agent does not need.

Strong style priority:
Make the result as daily, conversational, and meeting-like as possible. This is very important. Prefer natural back-and-forth dialogue over polished written summaries. It should feel like people are talking in a sync meeting, clarifying points, agreeing, interrupting lightly, confirming details, and deciding how to record the work.

Do not rewrite ordinary documents. Files without speaker/content style dialogue are not meeting rewrite targets. Docs/docx/wiki/tutorial/onboarding/excerpt/readme resources, document summaries, internal documentation snippets, API instructions, benchmark artifacts, and reference material are not meeting files, even if they are useful context.

Very important constraints:

- Only edit files under the current working directory.

- You may edit the `messages` array itself when needed: split turns, merge turns, reorder nearby turns, add short natural acknowledgements, and adjust speaker distribution.

- Keep each message item in speaker/content style.

- For every meeting file you rewrite, delete all top-level fields except `messages`.

- Important non-message information must be moved into natural dialogue before the original field is deleted.

- Unimportant, redundant, or purely administrative non-message information should be deleted without forcing it into dialogue.

- For nested structured fields outside `messages`, convert important values into realistic back-and-forth conversation instead of preserving them as schema keys or dumping them as lists.

- The final rewritten meeting JSON must have exactly this top-level shape: one object with one key named `messages`.

- Do not edit non-candidate files.

- Do not edit files outside this directory.

- Do not edit middleware config files such as `middleware/registry.json`, `middleware/hidden\_context.json`, or `middleware/agent\_config.json`.

- Do not edit docs, doc excerpts, onboarding docs, wiki excerpts, tutorial notes, non-meeting data files, experiment outputs, benchmark rows, schemas, manifests, code configs, or pure artifact JSON.

- Do not turn a document excerpt into a conversation. If the source is a document, leave it as a document.

- If no candidate has explicit speaker/content-style dialogue, leave every file unchanged.

- Preserve valid JSON. The file must still parse as JSON after editing.

- Do not preserve the surrounding top-level JSON schema for meeting files. Keeping only `messages` is required.

- Do not change factual content: names, dates, numbers, task decisions, model names, file paths, URLs, project names, conclusions, TODOs, blockers, and assignments must remain the same.

- Do not invent new facts, decisions, people, claims, or results.

- Do not delete useful information.

- The goal is style only: less stiff, less synthetic, more like real Chinese meeting speech.

Target style:
- Natural daily spoken Chinese, with realistic starts/stops, short fragments, mild repetition, and informal connectors.

- * Strongly prefer realistic conversation or meeting-note dialogues: short turns, acknowledgements, follow-up questions, clarifications, "yes, yes", "hmm", "OK", "right?", "let's leave it here for now", and similar natural meeting flows.

- It is better to split a stiff long paragraph into several natural turns when that makes the exchange feel like an actual meeting.

- * It can use phrases such as "it's just", "I think", "yes, yes", "right?", "then", "maybe", "let's leave it for now", "we can come back to it later", "how about", "this thing", "more or less", "not really workable", and "we can give it a try".

- It should feel like people are talking in a meeting, not like a formal report.

- Do not overdo filler words. It should be human, not parody.

- Keep technical terms intact: API names, model names, dataset names, benchmark names, file names, commands, links, and numbers should stay recognizable.

Style reference as JSON, do not copy facts from this reference:
\{style\_reference\}

Few-shot examples:
\{fewshot\_examples\}

Now inspect the candidate files and make the edits directly. If there is no meeting-like JSON in this directory, leave every file unchanged.
\end{promptbox}

\subsection{Instruction Alignment Loop}
\label{sec:app_prompt_alignment}

After an agent task is reconstructed and executed, an alignment checker determines whether the execution corresponds to the work implied by the original document diff. If the checker finds a mismatch, a reviser rewrites the agent task while preserving the intended work semantics.

\begin{promptbox}{Execution alignment prompt}
You will receive:

- The original worklog markdown diff

- The inferred work from pipe2

- The agent task constructed by pipe2

- The final message and major file changes after the agent's actual execution

Your task is to determine whether this agent task and its actual execution trajectory are performing the type of work reflected in the original worklog diff.

Judgment rules:

- Focus only on whether the actions performed are correct, not on whether the specific content matches the markdown diff.

- If the worklog is about research/verification/analysis, and the agent is also conducting related research/verification/analysis, even if the research conclusions, material sources, and details are inconsistent with the markdown diff, it should still be judged as `action\_aligned`.

- If the worklog is about accumulating prompts/rules/plans/summaries, and the agent is also producing similar specifications/plans/summaries, even if the text content is different, it should still be judged as `action\_aligned`.

- Do not predefine limited types for worklogs, and do not rely on fixed keyword enumeration. You need to first analyze the worklog using open text: what specific content was added/modified; what actions did this content imply the executor actually perform; which are definitive results, and which are pending tasks or process logs.

- Then deduce `expected\_agent\_action`: what the agent should actually do if this worklog were to be reconstructed into an agent task. `expected\_agent\_action` must be open text and can be something like "ensure X runs successfully/attempt to start and verify X/download and check X/complete and run X/organize the prompt word framework for X," etc., not selected from fixed categories.

- If the worklog contains definitive execution results, such as something running successfully, completed, downloaded, verified, fixed, or implemented, you should usually deduce an action that can accomplish this, rather than simply having the agent confirm whether this event exists.

- "Ensure/attempt to run X successfully" and "confirm whether X has run successfully" are not the same task: the former requires the agent to attempt to start, configure, invoke, and verify X, and record observable evidence or reasons for blocking; the latter only audits the current state. When the worklog shows that X has been successfully executed, the revision should change the instruction to a specific execution task related to ensuring X's success, rather than simply making a yes/no judgment.

- If the agent\_task weakens, shifts, or changes the expected\_agent\_action, for example, changing "ensure X is successfully executed" to "confirm whether X has been successfully executed," it should be judged as `action\_mismatch`, and the issue\_types should include `state\_presupposition\_mismatch` or `wrong\_action\_type`.

- Only when the agent\_task or execution trajectory has clearly changed to a different type of work should it be judged as `action\_mismatch`. For example: originally investigating data filtering rules, but instead implementing an unrelated webpage; originally organizing plans, but instead modifying business code; originally verifying resources, but only writing an unrelated summary.

- Do not directly judge an action as wrong simply because the agent's final conclusion contradicts the conclusion in the worklog diff; this may simply be due to different results from a re-investigation. Only when such a contradictory conclusion indicates that the agent was asked to perform a different type of task that it shouldn't have been doing should it be considered supporting evidence.

- If the evidence is insufficient to make a judgment, return `needs\_review`; do not force a judgment of aligned or mismatch.

- Provide a structured `revise\_note` and a specific, actionable `suggested\_reconstruct\_guidance` to reconstruct the `agent\_task`.

- The `revise\_note` should contain seven types of information:

1. `worklog\_observation`: What definite content, process records, or pending signals are in the worklog.

2. `expected\_agent\_action`: What the agent should actually perform, deduced from the worklog. This must be open text; do not use type enumeration.

3. `instruction\_gap`: The difference between the current `agent\_task` and `expected\_agent\_action`, such as weakening, offsetting, changing the task, or auditing without execution.

4. `instruction\_problem`: Where the current instruction is wrong; point out the task structure that was incorrectly rewritten, rather than criticizing the specific execution result.

5. `preserve\_work\_semantics`: Which work themes, task stages, or action relationships should be preserved after the correction.

6. `avoid\_instruction\_patterns`: Which instruction patterns should be avoided after the revision.

7. `rewrite\_strategy`: How to rewrite the instruction into a new, executable one.

- `suggested\_reconstruct\_guidance` should be a natural language compressed version of `revise\_note`, containing three types of information:

1. What is `expected\_agent\_action`?

2. What are the differences between the current instruction and `expected\_agent\_action`?

3. How should it be rewritten, and which instruction patterns should be avoided?

- Neither `revise\_note` nor `suggested\_reconstruct\_guidance` should hardcode specific status conclusions in the worklog as mandatory inherited facts; avoid writing "X must be retained as successful / Y as verified". Instead, write "The worklog implies that the agent task should ensure/attempt to complete X, and log the reason for the blockage if it fails." - If the problem is `state\_presupposition\_mismatch`, the guidance should explicitly state:

- The current error might be that a task of "ensuring/attempting to complete an action" has been changed into a yes/no audit task of "confirming whether it has been completed".

- The instruction should be changed to an executable task surrounding `expected\_agent\_action`, such as starting, configuring, calling, verifying, downloading, running, modifying, completing, and cleaning up artifacts.

- If the current environment cannot complete the task, the agent should also log the actions attempted, the specific bottlenecks, and the subsequent required conditions, rather than treating the failed audit conclusion as the main artifact.

- The agent can be allowed to log "current environment cannot be verified/verification failed/environment gap exists," but this should only be as supplementary information and not the primary task objective.

- The `revise\_note` should be specific enough to directly guide the reviser, but should not write complete instructions for the reviser; the complete instruction should be reshaped by the reviser in conjunction with the current `agent\_task`.

- Guidance should be written as a direct suggestion for modification that can be used directly by the reviser, avoiding overly abstract statements.

Returns STRICT JSON, returning only a single JSON object.
\end{promptbox}

\begin{promptbox}{Instruction revision prompt}
You will receive:

- The inferred work already deduced by pipe2

- The currently reconstructed agent\_task

- The LLM checker's assessment and modification suggestions for agent\_task/execution alignment

Your task is: to reshape agent\_task based on the checker's modification suggestions without modifying the old reconstruction prompt process.

Requirements:

- Use alignment\_check.revise\_note first; if no revise\_note is available, then use suggested\_reconstruct\_guidance.

- Expected\_agent\_action in revise\_note is the most important reconstruction target; instruction\_gap/instruction\_problem are used to locate errors in the current task structure; preserve\_work\_semantics are used to preserve task phases, work topics, and action relationships; avoid\_instruction\_patterns are used to delete or rewrite dangerous expressions; rewrite\_strategy is used to generate new instruction mainlines.

- Only correct the task semantic issues pointed out by the checker; do not replicate specific text to conform to markdown diffs.

- If the checker points out that task prerequisites have been rewritten, for example, changing "ensure/attempt to run, start, verify, download, fix, achieve something" to a yes/no audit of "whether it runs/whether it is completed/whether it is available," you should revise the task structure: Have the agent actually execute, verify, and log evidence or reasons for blockages around the expected\_agent\_action, rather than simply re-evaluating whether the status is met.

- Do not interpret checker guidance as requiring the inheritance of every specific factual conclusion in the worklog. The focus should be on preserving task phases, work topics, and action relationships.

- Instructions should describe "what the agent should do next," not how the final documentation should be written.

- Keep tasks as high-level, executable handovers; do not write them as step-by-step SOPs.

- Retain reasonable success criteria, outputs, and caveats in the current agent\_task, but simultaneously revise any statements that conflict with the checker's opinions.

- If the checker determines it is not an action mismatch, make as few changes as possible or revert to the original task.

- Do not introduce new resources, interfaces, file paths, or specific facts that the checker has no basis for. - Instructions can be allowed to require the agent to log "current environment cannot be verified/verification failed/environmental gap exists," but this should only be used as supplementary information and cannot be the primary objective.

Return STRICT JSON, only one JSON object is returned, the schema is the same as reconstructed\_agent\_task.json:
\{ 

    "subtask\_id": "string", 

    "status": "ok | needs\_review | insufficient\_evidence", 

    "agent\_task": \{ 

        "title": "string", 

        "instruction": "string", 

        "resource\_dependencies": \{ 

        "static\_files": [\{"path": "string", "why\_needed": "string"\}], 

        "middleware\_operations": [\{"name": "string", "purpose": "string", "usage": "string"\}] 

        \}, 

        "success\_criteria": ["string"], 

        "expected\_outputs": ["string"], 

        "caveats": ["string"] 

    \}
\}
\end{promptbox}

\subsection{Controlled Trajectory Rewrite}
\label{sec:app_prompt_trajectory_rewrite}

After obtaining a semantically aligned execution, we rewrite the trajectory so that the message log and workspace evolution are natural, executable, and consistent with the validated target document state.

\begin{promptbox}{Trajectory rewrite prompt}
You are a trajectory rewrite planner. Your task is to reconstruct the real target document diff into a natural, executable, and disk-persistent agent message log.

You will receive:

- Original agent task

- Real target document before/after markdown

- Behavioral summary of the original message log

- Summary of changed files in the workspace after execution

- Summary of the current workspace file tree and candidate target files

The objective must satisfy three constraints:

1. The rewritten trajectory must execute the task from scratch like the real agent, without appearing to know the final document answer in advance.

2. There must be a natural target file write action at the end, and the written content must be verbatim equal to the after\_markdown.

3. Before writing the target file, sufficient evidence must have naturally generated in the trajectory; the workspace must also have corresponding supporting files or original changed files.

Important requirements:

- Do not use phrases that reveal the benchmark, such as "based on the real diff," "target answer," "not written incorrectly," or "verbial matching."

- Do not hardcode the task type. Plan reasonable execution steps based on the current target document diff.

- Refer to the original message\_log pattern: for example, first check the workspace, then query/read files/call APIs/write files/validate/summarize. Do not copy irrelevant content.

- If the target document records research, the trajectory should first show research actions; if it records running/fixing/implementation, the trajectory should first show attempts, verifications, or blocking records; if it records plans/summaries, the trajectory should first show information organization and summarization.

- Supporting files can be generated, but they must be files that the agent will naturally write during execution. The target document should be the final brief record, not the only output.

- The execution\_plan should only record the process of "generating evidence/completing the task," do not write the final target\_path in the execution\_plan; the final target file will be appended uniformly by the outer program.

- When planning, first work backward from after\_markdown: if this task is executed from scratch, what checks, research, runs, verifications, and organization are needed to naturally obtain these records? Then express this using the interaction rhythm of the original message\_log.

- `target\_path` must be determined based on the current workspace file tree, `candidate\_target\_files`, `doc\_title/doc\_id`, `before\_markdown` content similarity, and task semantics. Do not default to `worklog/worklog.md`; only select it if current evidence indicates it is the target file.

- Output must be STRICT JSON.

Return schema:

\{
    "status": "ok" | "needs\_review",

    "execution\_plan": [

        \{
            "purpose": "Natural language explanation of why this stage is necessary",

            "assistant\_message": "Agent's explanation of the next steps to the user/itself",

            "tool\_command": "Optional; shell command or equivalent action",

            "tool\_output": "Optional; summary output that the command should reasonably see",

            "workspace\_writes": [

            \{"path": "Path relative to /workspace", "content": "Complete file content"\}

            ]

        \}

    ],

    "target\_path": "Target file path relative to /workspace",

    "target\_path\_reason": "Why this file was chosen as the final write target",

    "target\_content": "Must be verbatim equal to after\_markdown",

    "worklog\_path": "Legacy-compatible, optional",

    "worklog\_content": "Legacy-compatible, optional",

    "worklog\_line\_support": [ \{"line": "A line in the target document", "supporting\_plan\_indices": [0, 1], "supporting\_paths": ["..."]\}

    ],

    "notes": ["Any points requiring human attention"]

\}
\end{promptbox}

\subsection{Evaluation Prompts}
\label{app_b}

The following boxes reproduce the prompts used to construct and evaluate the reported metrics. Task-specific inputs are supplied at runtime; the boxes retain the evaluation instructions and required output schemas.

\subsubsection{Downstream Evaluation Prompts}

\begin{promptbox}{Workspace Score}
You are a strict workspace evaluator for one Membench subtask.

Module rubric:

`workspace\_task\_completion`

Question:

Did the final Docker workspace actually complete the task implied by the worklog/document change and the task instruction?

Use evidence from:

- `./input/task\_context.json`, especially `task\_description`, `metric\_description`, `planned\_tasks`, `inferred\_work`, and `agent\_task`.

- `./input/worklog\_ground\_truth.json`, especially `worklog\_changes[*].before\_markdown` and `worklog\_changes[*].after\_markdown`.

- `./input/agent\_trace.json`, optional execution evidence from the evaluated run.

- `./input/rollout.jsonl`, optional supporting raw evidence.

- `bundle.docker.candidate\_image\_ref`.

- `bundle.docker.reference\_image\_ref`.

Core workflow:

1. Read the task context and reference document change to understand the actual task.

   - Identify the task type: research, analysis, code/config/environment work, prompt/rule/pipeline work, data work, or mixed.

   - Identify the core deliverable and the quality signal the workspace should contain.

   - Do not infer requirements from filenames alone.

2. Inspect `bundle.docker.reference\_image\_ref` first.

   - Treat it as the solid reference baseline for this task, roughly around the 80-point level.

   - It is not perfect and does not need to be copied.

   - Extract what it does well as content/task-quality requirements, not as path or filename requirements.

3. Write `./output/criteria.json` before inspecting the candidate.

   - `minimum\_requirements`: what a barely passing workspace must accomplish for this task.

   - `reference\_requirements`: what the reference workspace accomplishes at a solid level.

   - `better\_requirements`: what would be meaningfully better than the reference workspace.

   - Good requirements describe task quality: correctness, completeness, depth, breadth, concrete evidence, validation, robustness, usefulness, and actionability.

   - Bad requirements overfit to filenames, paths, exact document names, or exact wording.

   - Write enough requirements to cover the task in detail. Small tasks may need 5-8 requirements; broader research/code/analysis tasks often need 12-20 requirements.

   - Use the same requirement ids across `minimum\_requirements`, `reference\_requirements`, and `better\_requirements`, so each basic/reference/better standard is easy to compare.

- Requirements should be specific and checkable. Use task-relevant concrete standards such as scope, coverage, behavior, correctness, validation, environment setup, edge cases, evidence quality, or actionability.

- Numbers or ranges are only one way to make a requirement concrete. Use them when they naturally fit the task, but do not force numeric requirements when behavior, correctness, or validation is the real standard.

  - Example: if the reference clearly inspected about 30 relevant webpages/sources for a research task, a concrete requirement can say the reference-level workspace should cover about 30 relevant sources with useful categorization; a basic passing workspace might cover at least 15 relevant sources; a better-than-reference workspace might cover more sources or provide stronger validation, comparison, and risk analysis.

4. Inspect `bundle.docker.candidate\_image\_ref`.

   - Find what the candidate actually produced or changed.

   - File paths are evidence only; they do not define success unless the task specifically requires a path.

5. Directly compare the candidate workspace against the reference workspace.

   - Compare content and task quality, not whether the same file name exists.

   - Ask whether the candidate is below minimum, near minimum, weaker than reference, close to reference, or better than reference.

6. Score the candidate using all three anchors: task requirements, the minimum requirements, and the direct reference comparison.

Docker inspection:

- Use Docker commands to inspect `/workspace` in both images.

- Read relevant files, not just filenames.

- Inspect code, scripts, configs, generated artifacts, documents, and environment setup when relevant.

- Run substantive validation commands when the workspace contains runnable code and the task requires running, fixing, validating, downloading, building, testing, or checking behavior.

- Do not rely only on a candidate's written report, final message, or claimed command output. Enter the Docker image and execute the relevant workflow yourself when it is possible: scripts, tests, builds, package commands, config loading, server startup, tool/client calls, data-processing pipelines, generated-artifact checks, or other task-specific verification.

- Validation is not limited to a smoke test. If the task needs a complex command or multi-step run to know whether the workspace works, run that command or workflow.

- If the full workflow requires unavailable external services, GPU/display hardware, credentials, network access, or would take an unreasonable amount of time, run the strongest feasible partial validation and explain the exact blocker.

- When you run a validation command, cite the observed result in the requirement `reason`: what command or behavior you checked, whether it passed, and where it failed if blocked.

- Do not modify either image. Use `docker run --rm` for inspection.

Scoring logic:

- The reference image gives the solid baseline for this exact task.

- The minimum requirements define what just barely counts as basic completion.

- The better requirements define what would be meaningfully better than the reference baseline.

- Do not score only by checking the minimum requirements. Direct comparison with the reference workspace is a core part of this metric.

- The reference comparison should shape the requirements themselves. If the reference succeeds because it researches more sources, gives clearer categorization, validates behavior, or implements a more complete solution, write that as a reference requirement.

- Better requirements should be plausible improvements over the reference, not generic ideals. They should say what deeper, broader, more validated, more robust, or more useful work would look like for this exact task.

- Compare the candidate and reference on task-specific content quality, depth, breadth, correctness, implementation completeness, validation evidence, robustness, and usefulness.

- For research tasks, compare whether the candidate covers the same conceptual ground as the reference, and whether it is deeper, broader, better organized, more concrete, or more actionable. Do not require the same filenames.

- For code/config tasks, compare whether the candidate has the same functional behavior and validation quality as the reference, not whether it copied the same files line by line.

- For toolchain/code/environment tasks, runnable validation is part of the task quality. A script, config, install, or report is not enough by itself if the task requires something to run; the requirements should ask for inspected execution evidence.

- For analysis tasks, compare whether the candidate is as grounded, complete, nuanced, and useful as the reference.

- Candidate below the minimum requirements should usually score below 60.

- Candidate that only meets the minimum requirements should score around 60.

- Candidate clearly better than the minimum but weaker than the reference baseline should usually score 60-79.

- Candidate close to the reference baseline should score around 80.

- Candidate satisfying important better requirements can score above 80.

- Score above 90 only when the candidate has strong task-specific correctness, depth, validation, coverage, robustness, or usefulness beyond the reference baseline.

Anti-hack review:

Penalize heavily for:

- Only writing a worklog/report that claims success while required code/config/artifacts are missing.

- Empty stubs, placeholder files, toy examples, or fake pass outputs.

- Hardcoded outputs that avoid the task's real requirement.

- Removing or weakening checks/tests instead of fixing the underlying issue.

- Creating unrelated polished artifacts while the required task implementation is absent.

- Claiming a command/test/build passed without inspected workspace evidence.

- Overfitting to the worklog text instead of doing the task.

Criteria file:

Before returning the final result, write exactly this JSON object to `./output/criteria.json`:

\{

  "task": "short task summary",

  "candidate\_image": "candidate docker tag",

  "reference\_image": "reference docker tag",

  "minimum\_requirements": [

    \{

      "id": "req\_1",

      "requirement": "detailed minimum requirement for basic completion"

    \}

  ],

  "reference\_requirements": [

    \{

      "id": "req\_1",

      "requirement": "what the reference image shows for the same requirement"

    \}

  ],

  "better\_requirements": [

    \{

      "id": "req\_1",

      "requirement": "what would be meaningfully better than the reference for the same requirement"

    \}

  ]

\}

Requirement writing rules:

- Requirements must describe task/content quality, not exact file names or paths.

- Do not write requirements like "the workspace contains /workspace/foo.md" or "uses the same filename as reference".

- File paths and filenames are evidence only. Put them in `candidate\_files` or in the `reason` field when judging a requirement.

- Requirements should be concrete, but only in ways a different valid solution can satisfy.

- Write enough requirements to make scoring discriminative. For broad tasks, prefer many concrete requirements over a few vague ones; around 20 requirements is acceptable when the reference workspace has enough substance.

- Keep each requirement focused on one checkable quality point. Do not merge unrelated quality points into one vague requirement.

- Use matching ids across `minimum\_requirements`, `reference\_requirements`, and `better\_requirements`; for example, `req\_7` should describe the same aspect at basic, reference, and better-than-reference levels.

- Good concrete requirements include task-specific scope, coverage, behavior, correctness, validation, dependency/config correctness, environment reproducibility, edge cases, and evidence quality.

- Numeric thresholds are useful only when the task naturally has a countable scope. They are examples of concreteness, not the main goal.

- If the reference suggests a useful count or range, use it. If the important standard is behavior, correctness, validation, or environment quality, write that standard concretely instead.

- Bad concrete requirements require identical filenames, identical wording, identical line count, or exact document layout unless the task explicitly requires that.

- For research tasks, requirements should describe quality such as depth, breadth, correct categorization, concrete examples, useful comparisons, limitations, and actionability.

- For code tasks, requirements should describe behavior, integration, correctness, validation, robustness, and maintainability.

- For environment/config tasks, requirements should describe dependency correctness, runnable commands, path consistency, configuration validity, and reproducibility.

- For toolchain, code, script, environment, or "run/fix/validate" tasks, include explicit requirements for actual execution or the closest possible safe validation. Do not let a candidate score high only because it wrote plausible code, commands, or a report.

- Validation requirements should exist at all three levels:

  - Minimum: the candidate must run the core command/script/tool or explain a real inspected blocker with evidence.

  - Reference: the candidate should provide execution evidence comparable to the reference workspace, such as command output, installed artifacts, generated files, test results, logs, or inspected failure location.

  - Better: the candidate should provide stronger end-to-end validation, preserved logs/scripts, generated artifacts, repeated checks, edge-case checks, or clear reproducibility.

- If a task says something was "run through", "fixed", "verified", "connected", "validated", "started", "built", "downloaded", or "tested", write a requirement that checks whether this actually happened in the workspace. Treat unverified claims as weak evidence.

- If the candidate includes runnable code, a toolchain, scripts, tests, generated artifacts, or environment setup, verify the important executable paths yourself. This may require complex commands or multi-step workflows. If you cannot run them, explain the concrete blocker and do not give the same credit as a verified run.

- For analysis tasks, requirements should describe coverage, grounding, nuance, alternatives, failure modes, and useful conclusions.

- `minimum\_requirements` are the basic passing requirements.

- `reference\_requirements` are stronger requirements inferred from the reference image.

- `better\_requirements` describe what would be deeper, broader, more detailed, better validated, more robust, or more useful than the reference.

Concrete requirement examples:

- Research task where breadth matters:

  - Minimum: cover the main relevant source categories and record why each category matters.

  - Reference: cover the important source categories with concrete examples, correct categorization, useful comparisons, and limitations.

  - Better: add stronger coverage, validation notes, collection feasibility, risk analysis, or more actionable prioritization than the reference.

- Research task where the reference clearly has a countable scope:

  - Minimum: if the reference inspected about 30 relevant webpages/sources, a basic completion might inspect at least 15 relevant sources and record the main findings.

  - Reference: cover about 30 relevant sources with categories, examples, and useful comparisons matching the inspected reference depth.

  - Better: cover more than the reference or add materially better validation, source-quality checks, risk analysis, prioritization, or actionability.

- Research task where the reference distinguishes core, adjacent, and weak sources:

  - Minimum: separate clearly relevant sources from adjacent or weak sources.

  - Reference: explain why each category matters and give concrete caveats for adjacent sources.

  - Better: add an actionable prioritization plan and criteria for accepting or rejecting future sources.

- Code task:

  - Minimum: implement the core requested behavior in the relevant code path, avoid obvious stubs, and run at least one focused command/test/manual check that exercises the core behavior.

  - Reference: integrate the behavior cleanly, update related config or tests, and preserve inspected validation evidence comparable to the reference.

  - Better: handle important edge cases, add meaningful tests, preserve reproducible logs or commands, and keep the implementation maintainable.

- Environment/config task:

  - Minimum: required dependencies/config files exist and the main command is actually tried, or a real blocker is inspected and documented with evidence.

  - Reference: commands, paths, and dependencies are consistent and there is inspected evidence of a successful or partially successful run, including where execution stops if blocked.

  - Better: provide reproducible setup, preserved logs/scripts, clear failure handling, validation for the expected runtime, and generated artifacts when the task requires them.

- Toolchain validation task:

  - Minimum: verify the requested toolchain component by running the relevant command or client path far enough to prove the installed component is real, and state the exact blocker if full completion is impossible.

  - Reference: match the reference's evidence level: package/version checks, startup or handshake evidence, inspected entry points, command output, downloaded/generated artifacts, and a conservative readiness judgment.

  - Better: complete an end-to-end run if possible; otherwise preserve raw logs/scripts and produce the smallest meaningful generated output or failed call evidence that proves where the chain stops.

- Analysis task:

  - Minimum: answer the core question with evidence from inspected materials.

  - Reference: cover main alternatives, limitations, tradeoffs, and concrete examples.

  - Better: add deeper failure-mode analysis, clearer decision criteria, and more actionable next steps.

Return exactly this JSON object:

\{

  "module": "workspace\_task\_completion",

  "score": 0,

  "task": "short task summary",

  "criteria\_file": "output/criteria.json",

  "candidate\_image": "candidate docker tag",

  "reference\_image": "reference docker tag",

  "candidate\_files": [

    \{"path": "path inside /workspace", "summary": "what is actually present"\}

  ],

  "minimum\_requirements": [

    \{"id": "req\_1", "requirement": "requirement from criteria.json", "result": "short judgment", "reason": "short evidence-based reason"\}

  ],

  "reference\_requirements": [

    \{"id": "req\_1", "requirement": "requirement from criteria.json", "result": "short judgment", "reason": "short evidence-based reason"\}

  ],

  "better\_requirements": [

    \{"id": "req\_1", "requirement": "requirement from criteria.json", "result": "short judgment", "reason": "short evidence-based reason"\}

  ],

  "score\_rationale": "short concrete explanation comparing candidate against minimum requirements and the reference workspace",

  "confidence": 0.0,

  "summary": "short concrete judgment"

\}
\end{promptbox}

\begin{promptbox}{Preference Rubric Construction}
You are the global preference extractor for Membench.

Objective:

Use only this participant's real message logs and the corresponding Markdown before/after changes in `doc\_change.json`. From this evidence, extract one globally reusable set of preference metrics. The same global metrics will later be used to evaluate every subtask for this participant.

You must complete two tasks:

1. Identify the participant's primary document: the path of the main, persistent worklog or work record.

2. Derive approximately 30 global preferences from all reference Markdown changes. Each preference must state a clear direction, not merely name a dimension.

Input files:

- `./input/reference\_subtasks.json`

  - Lists all subtasks.

  - Each subtask contains task information, root preference metrics, and `before\_markdown`/`after\_markdown` from `doc\_change`.

  - `doc\_change` contains Markdown document edits produced by the reference run. Consider only Markdown/document records, not code, scripts, webpages, packages, or asset files.

- `./input/message\_logs/\textless{}subtask\_id\textgreater{}.json`

  - Contains the real message log for each subtask.

  - Use it to determine which Markdown worklogs, reports, specifications, prompts, or planning documents the reference run actually wrote, and to identify the primary document path.

Primary-document identification requires special care:

- Do not extract it using regexes, fixed keywords, or filename rules. Filenames, titles, paths, and write behavior are clues, but the final judgment must come from reading the actual message logs and document changes.

- Review all Markdown/document changes before choosing the primary document. Do not rely only on the first subtask or one filename.

- The primary document is the participant's long-term durable record. It typically carries updates across multiple subtasks rather than serving as the complete deliverable for one task.

- Names such as `xxx worklog.md`, `work log.md`, `daily.md`, `weekly.md`, `record.md`, and `log.md` are clues, not rules. A primary document may have an unusual name, and an artifact may also contain `log` in its name.

- Its content usually resembles an ongoing record: dates, times, tasks, TODOs, tentative checks, attempts, results, failures, next steps, short bullets, work-session order, participant names, task titles, and incrementally appended passages.

- It is usually updated by appending, inserting near related context, adding the current task record, or extending an existing heading or paragraph, rather than by creating a new self-contained report.

- It may cover many topics without fully developing each one. Its purpose is to support later continuation, not to become a polished one-off document.

- A document is more likely to be primary if it is repeatedly read and appended across subtasks, repeatedly mentioned as the final record, or explicitly called a worklog or record in the message logs.

- If `after\_markdown` resembles a personal worklog title, a daily record, a task list, chronological notes, or short work items but does not contain a path, use the message log to locate the actual path.

- If several documents look like worklogs, choose the one that most clearly serves as the participant's persistent primary record. Do not select a longer or more polished artifact merely because it is more complete for one task.

Distinguish supporting documents carefully:

- Other documents may be research reports, rules, prompts, plans, designs, explanations, checklists, experiment logs, generated reference documents, or doc-id-like Markdown files.

- They usually act as task-specific artifacts: they are more complete, formally titled, independently readable, or limited to one topic.

- These documents may be important, but they are not the primary document. When a reference change targets one of them, later evaluation should place less weight on matching its path and more on content, structure, language, scope, evidence, and preference consistency.

- If a reference run updates both the primary document and a supporting document, interpret their roles separately: the primary document usually records the final work state, while the supporting document provides fuller material or explanation.

- If a task updates only a supporting document, do not redefine that document as primary. The primary document is a global concept, not simply the file modified in the current subtask.

The primary document path is required:

- Always return `main\_worklog\_path`, which downstream single-preference evaluation will use directly.

- It must be a workspace path inferred from the real message logs and document changes.

- If it cannot be determined, return `main\_worklog\_path: ""` and explain why in `summary`. Never invent a path.

- `main\_document.reason` must give one concise reason, such as repeated worklog appends, use as the final record across tasks, or explicit identification as a work record in the message logs.

- `main\_document.changed\_subtasks` must contain only subtasks that clearly updated the primary document, not every subtask that read it.

From a dimension to a preference:

- A dimension is only an observation axis, such as length, sentence length, table use, append versus rewrite, or preservation of failed attempts.

- A preference must reflect the aggregate pattern across all subtasks, not an arbitrary conclusion from one subtask.

- For each dimension, compare all reference Markdown changes. Determine the dominant direction, the exceptions, and whether the pattern is stable enough to be global.

- If most changes are short records with little explanation and lightweight bullets, while only a few are long reports, derive a preference for concise updates. Treat the long artifacts as contextual exceptions.

- If most short records avoid tables but a few formal comparison documents use them, state that the primary document and short records favor non-tabular presentation, while artifact documents may use tables when comparison requires them.

- If most changes append small updates near existing context while a few create complete standalone documents, state that the primary document favors local append/merge edits while artifacts may be written independently.

- If a dimension has no clear direction and the contexts do not explain the split, do not force a strong preference. Choose a more stable dimension.

- Every preference must state the favored direction. Examples:

  - Dimension: word/character budget. Preference: favor short updates close to the reference record length; do not expand a brief worklog entry into a full report.

  - Dimension: sentence length. Preference: favor short, note-like sentences over long explanatory paragraphs.

  - Dimension: append versus rewrite. Preference: favor small additions or merges near the existing context over replacing a local update with a new complete document.

  - Dimension: uncertainty. Preference: preserve states such as still investigating, pending verification, uncertain, or try first rather than presenting preliminary observations as final conclusions.

- The output must contain preferences, not dimension names alone.

Requirements for global metrics:

- Produce approximately 30 metrics; 27--33 is acceptable. Do not retain duplicate, overly narrow, or weakly supported preferences merely to reach a count.

- Every metric must be globally reusable. Do not mention a specific subtask, API, URL, project, command, date, or filename.

- Specific evidence may inform the trend, but the output preference itself must remain general and must not include long evidence passages.

- Every metric must have a direction: short versus long, append versus rewrite, tentative judgment versus formal conclusion, table versus non-tabular form, and so on.

- Keep metrics diverse. Do not restate the same idea in different words.

- If two metrics would normally receive the same score from the same evidence, merge them or remove one and replace it with a different dimension.

- Root preferences are references only. Do not simply paraphrase them; derive richer global preferences from all reference edits.

The review of global metrics is the most detailed and important part of this process:

- Work incrementally rather than generating the entire list at once.

- First create `./output/global\_preferences\_draft.json` and write an initial batch of 8--12 preferences. Review that batch, delete or revise invalid items, add a small number from different dimensions, review again, and continue until the list reaches approximately 30.

- Review every batch before adding the next. Do not draft everything first and perform only one superficial final review.

- Whenever a review removes a duplicate, non-global, or directionless preference, immediately replace it with a preference from a different dimension and review the replacement.

- Write the stable version to `./output/result.json` and return the same JSON object to stdout. `./output/global\_preferences\_draft.json` is temporary and need not remain. Do not include discarded drafts in the final result.

- Perform multiple rounds of drafting, review, revision, and completion until the list is stable.

- In every round, recheck global applicability, direction, overlap with root preferences, overlap with retained preferences, and whether message-log/document evidence can actually support evaluation.

- Return only preferences that survive all review rounds.

For every generated preference, check global applicability:

- Would it still be evaluable if the subtask, technical domain, and filename changed?

- If not, rewrite it more generally or remove it.

- Do not write "preserve the Three.js demo link"; write "favor retaining concrete, reviewable sources or examples without requiring exact names."

- Do not write "update a particular person's worklog file"; write "when the reference updates the primary document, favor writing the final record back to that document."

For every generated preference, check direction:

- Do not output only "word count" or "sentence length."

- State the favored side: shorter or longer, short or long sentences, append or rewrite, rough record or formal report, preserved uncertainty or final conclusion.

- If direction is uncertain, do not force a strong preference. Use a better-supported dimension.

For every generated preference, check overlap with root preferences:

- A root preference may inspire analysis, but a derived global preference cannot simply restate it.

- If a root preference already covers concision, do not add many variants of short, sparse, or unexpanded writing. Consider genuinely distinct dimensions such as line budget, heading count, append behavior, or scope boundaries, and remove them too if they rely on the same evidence.

- If a root preference already covers structure, do not accumulate redundant bullet, list, and section preferences.

- If it already covers informal work-record language, do not accumulate redundant preferences about light conversational tone, mixed technical language, or avoiding formal register.

- If it already covers concrete anchors, do not accumulate redundant preferences about retaining URLs, APIs, commands, and project names.

Compare every generated preference with previously retained `global\_preferences`:

- If two preferences mean nearly the same thing, remove the later one.

- If they would rely on the same passage, file behavior, or missing evidence, remove one.

- Do not retain several preferences that all penalize excessive length, excessive formality, failure to update the primary document, missing anchors, lost uncertainty, or needless structural complexity. Keep the clearest one and replace the rest with different dimensions.

- After every deletion, add a preference from a different dimension and review it again.

- Repeat deletion, revision, replacement, and review for multiple rounds.

Perform a final whole-list review:

- Avoid too many location-related metrics. A few may concern the primary document, but most of the list must not be about paths.

- Avoid filling the list with concise, short, sparse, or do-not-expand variants.

- Cover diverse dimensions: text form, editing behavior, the relationship between primary and supporting documents, record organization, expression granularity, evidence and verification, task scope, local language style, structural rhythm, and judgment or risk expression.

- Ensure every preference can be supported by message-log/document evidence visible to the single-preference evaluator.

- Keep the output concise; do not include full before/after text.

Few-shot dimension pool:

These are prompts for exploration, not fixed answers. Convert any selected dimension into a directional preference.

Text form:

- word/character budget; line budget; sentence length; paragraph length

- bullet count and nesting depth; heading count; table use

- blank-line and separator rhythm; balance of bullet lengths

- short work record versus long report

Information density:

- information per item; summary/detail ratio; number of examples

- amount of background explanation and repetition

- abstraction versus concrete record; scanability

- whether useful information is buried in low-value detail

Editing behavior:

- append versus rewrite; local patch versus whole-document restructuring

- preservation of prior context; restraint in deletion

- small traceable updates versus large untraceable rewrites

- insertion near related context versus an unrelated location

- durable record versus temporary analysis draft

- placement above or below existing text

- continuation of an existing topic rather than taking over the document opening

Primary/supporting-document relationship:

- use of the primary document as the durable record

- preservation of supporting documents in an auxiliary role

- clarity of the source of truth

- writing core conclusions back to the primary document

- avoiding use of an artifact as the only result

- different levels of detail in the primary and explanatory documents

- allowing fuller supporting documents without replacing the primary worklog

Record organization:

- chronological or work-session order; update granularity

- reuse of local labels and vocabulary; continuity with nearby context

- avoidance of unnecessary new classifications or taxonomies

- preservation of existing headings

- updating an existing section rather than creating excessive new sections

Expression granularity:

- observation versus conclusion; factual record versus recommendation

- process note versus final-outcome summary; raw note versus over-synthesis

- checklist versus narrative; label granularity

- omission discipline: do not expand everything the reference leaves implicit

Evidence and verification style:

- proximity of claims to evidence; source-attribution style; example-to-claim order

- visibility of failed attempts

- markers such as checked, not run, pending, or to verify

- explicit assumptions; preservation of negative evidence or rejected options

- preference for concrete records without requiring exact URLs, APIs, or commands

Task boundaries and scope:

- adherence to the requested small task; minimal sufficient deliverable

- avoidance of non-target work

- restraint in recording tool or process details

- avoidance of leaking excessive implementation detail when the deliverable is a record

- research depth proportional to the task

- avoidance of extra procedures that obscure the core preference

Local language style:

- modal language such as maybe, roughly, for now, still checking, try first, or later

- distinction among done, doing, and planned states

- imperative TODOs versus descriptive summaries; noun density

- lightweight work-record connectors versus formal-report connectors

- tolerance for abbreviations; reuse of local wording rather than full standardization

- preservation of mixed Chinese/English technical vocabulary when present

Structural rhythm:

- top-heavy versus bottom-heavy explanation; conclusion placement

- whether sections are forced into symmetry; heading specificity; hierarchy depth

- rough, scannable rhythm versus polishing into a specification

Judgment and risk expression:

- comparison style; ranking granularity; threshold or range expression

- trade-off wording; caveat placement; avoidance of unsupported certainty

- confidence gradients; visible but non-exaggerated risk

- preservation of preliminary findings as preliminary

Keep the output concise. Do not include long source passages, full before/after content, or input-file metadata.

Return exactly one JSON object and nothing else:

\{

  "module": "preference\_global\_derived\_metrics",

  "person": "person id",

  "main\_worklog\_path": "workspace path to the primary document, or an empty string if it cannot be determined",

  "main\_document": \{

    "path": "path to the primary document",

    "reason": "one sentence explaining why this is the primary document",

    "changed\_subtasks": ["subtask\_0001"]

  \},

  "global\_preferences": [

    \{

      "id": "global:d001",

      "dimension": "dimension name, for example word/character budget",

      "preference": "the participant's explicit, evaluable preference direction"

    \}

  ],

  "coverage": \{

    "subtask\_count": 0,

    "main\_document\_change\_count": 0,

    "other\_document\_change\_count": 0

  \},

  "confidence": 0.0,

  "summary": "one-sentence summary"

\}
\end{promptbox}

\begin{promptbox}{Preference Score}
You are a strict evaluator for one Membench preference subtask.

Module rubric:

`preference\_metric\_guarantee`

Question:

Did the evaluated run preserve this person's preference behavior for the current task?

Use evidence from the client input:

- `metric\_under\_test`: the one metric to judge in this API call.

- `task\_context`: task description, root preference metrics, and `main\_worklog\_path`.

- `worklog\_ground\_truth`: reference before/after markdown changes for this subtask.

- `reference\_document\_writes`: document writes extracted from the reference trace.

- `agent\_document\_writes`: document writes extracted from the evaluated run trace.

Document-write evidence:

- `reference\_document\_writes` is extracted from the reference agent execution trajectory and keeps only detected document files that were written by the reference run.

- `agent\_document\_writes` is extracted from the evaluated agent execution trajectory and keeps only detected document files that were written by the evaluated run.

- Read-only commands, code files, structured data files, and unrelated tool events are omitted.

- If a document-write list is empty, treat that as no detected document write in that execution trajectory.

Core rules:

- Do not derive new per-subtask preference metrics.

- Do not select, merge, split, or rewrite metrics. Judge only `metric\_under\_test`.

- Use `support: "ignored"` narrowly. Only ignore `metric\_under\_test` when it is completely outside this subtask's task type, reference markdown change, document role, writing behavior, and extracted write evidence, so no meaningful evidence could exist here.

- If `metric\_under\_test` is even somewhat related to the current reference change or evaluated run, do not ignore it. Score it as `none|weak|partial|strong` and explain the evidence or missing evidence.

- Use `main\_worklog\_path` to decide whether the reference change targets the person's primary durable worklog or a supporting artifact/document.

Important metric isolation rule:

Each API call judges exactly one preference: `metric\_under\_test`. Do not let any other preference, document-placement concern, root metric, global metric, or general worklog-quality concern affect the score unless it is explicitly part of `metric\_under\_test`. In particular, do not punish a run for not updating the primary worklog when the current metric is about style, tone, wording, concision, roughness, uncertainty language, evidence style, scope control, or another behavior that can be judged independently of where the text was written. For those behavior/style metrics, first ask: "Did the evaluated run show the behavior described by this metric in the relevant written output?" Judge that behavior directly. File path, main-worklog placement, artifact-vs-worklog choice, and durable-record target should matter only when the current metric itself is about placement, target document choice, durable record location, or preserving the document-reference change. If the metric is about writing style, then a supporting artifact, a secondary document, or another written note can still provide valid evidence of `strong` support when it matches the reference style. Conversely, a run can fail a placement metric while still strongly satisfying a style metric. Keep these judgments separate.

Important applicability rule:

Before choosing `none`, first decide whether the ground-truth/reference behavior actually gives evidence that `metric\_under\_test` applies to this subtask. The reference run is the anchor for applicability. If the current task and the ground-truth document change do not meaningfully exercise this preference, return `ignored`, not `none`. `none` is only appropriate when the metric is applicable because the reference/ground-truth behavior does show or require this preference, but the evaluated run clearly fails to show it. In other words: `ignored` means "this metric is not meaningfully testable in this subtask"; `none` means "this metric is meaningfully tested by the reference behavior, and the candidate completely misses it." Do not mark a metric `none` merely because the evaluated run lacks some behavior that the reference task never called for. For global preferences especially, many person-level habits will not apply to every subtask. If the GT/reference change is a short status note, do not force metrics about long artifact structure unless that behavior is actually relevant. If the GT/reference change is a placement update, do not force unrelated style or evidence metrics unless the reference also makes those behaviors observable. Always check the GT/reference first for applicability, then judge the candidate against the applicable metric.

Modified document summary:

- `gt\_modified` should list only the reference-changed markdown/document targets and a short summary.

- `candidate\_modified` should list only document-like files the evaluated run actually created or updated, especially markdown/worklog/report/spec/prompt/planning records.

- Do not list files that were only read, listed, or mentioned.

- Do not list non-document implementation artifacts unless they are the main written record being evaluated.

- Do not output raw before/after markdown. Summarize.

- Do not output runner metadata paths.

Support labels:

- `ignored`: the metric is completely unrelated to this subtask or impossible to judge from the available evidence.

- `none`: the metric applies and the evaluated run clearly fails it.

- `weak`: the metric applies but support is minimal, indirect, or mostly missing.

- `partial`: the evaluated run satisfies part of the metric but misses important details.

- `strong`: the evaluated run clearly satisfies the metric in the same preference behavior as the reference.

Judgment guidance:

- Judge the evaluated run against the specific behavior requested by `metric\_under\_test`.

- Use the reference before/after change and extracted reference document writes to understand the intended content, scope, tone, and level of detail for this subtask.

- For document-reference or document-placement metrics:

  - Decide whether the reference change targets the primary worklog or a supporting artifact/document.

  - If the metric is about durable-record placement, missing the primary worklog update is strong evidence of failure.

  - If the reference targets a supporting artifact/document, exact filename mismatch is usually less important than content, scope, tone, evidence, and preference behavior.

- For metrics not about document placement, do not over-focus on exact path, filename, main-worklog status, or whether the written output is in a supporting artifact. Judge only the preference behavior described by `metric\_under\_test`.

- For writing-style or tone metrics, compare whether the evaluated run uses the same level of formality, roughness, concision, language mix, status wording, and note-taking rhythm as the reference behavior.

- For scope and structure metrics, judge whether the evaluated run keeps the same task boundary, avoids unrelated expansion, and organizes the output in a way that matches the person's usual preference.

- For uncertainty or judgment metrics, check whether the evaluated run preserves tentative language, pending/blocked/failed states, comparative judgments, and limitations when the reference behavior calls for them.

- Content mismatch, missing central facts, wrong preference behavior, unsupported overreach, style mismatch, excessive verbosity, missing uncertainty, weak evidence, or poor scope control should lower the support label when relevant to the metric.

- Do not require byte-for-byte wording or exact URL/API/command names unless the metric is specifically about preserving exact concrete text. Prefer the same preference behavior and same recording style.

Metric application:

- `metric\_under\_test` may come from a human-selected root metric or a globally derived person-level preference. Apply it to this subtask when the current task gives meaningful evidence about that behavior.

- Relevance can come from document role, writing style, evidence-recording habits, status/uncertainty wording, scope control, concrete-detail preservation, output format, or the kind of artifact being produced.

- Use `ignored` only when the metric is unrelated to this subtask or the available evidence cannot meaningfully support or contradict it.

- If the metric is relevant but only weakly evidenced, choose `weak` or `partial` instead of hiding the uncertainty under `ignored`.

- Apply the metric in the current context. For example, a concision preference should not blindly penalize a task whose reference output is intentionally a longer artifact; judge whether the length and detail are appropriate for this subtask.

Return exactly this JSON object:

\{

  "metric\_id": "same id as metric\_under\_test.id",

  "support": "ignored|none|weak|partial|strong",

  "reason": "short reason with concrete evidence"

\}
\end{promptbox}

\begin{promptbox}{Preference Pairwise Comparison}
\# Preference Recall Compare Judge

Use this section only when the evaluation bundle has `module:

preference\_recall\_compare`.

You are comparing two evaluated runs for the same subtask:

- `candidate\_a`: one message log.

- `candidate\_b`: another message log.

Important:

- Candidate A and candidate B are randomly assigned by the runner.

- Do not assume candidate A is with-recall or without-recall.

- Do not infer labels from file names or directory names. The runner will map

  your raw candidate choice back to `with\_recall` or `without\_recall`.

Question:

Which candidate better preserves the user's preference/worklog behavior shown by

the reference before/after change and selected root preference metrics?

This is not a general task-quality contest. A long, polished artifact is not

automatically better. A candidate is better when it more closely follows the

user's durable-record and language-preference pattern for this subtask.

Evidence to use from the client input:

- `task\_context`, especially `task\_description`,

  `metric\_description`, `selected\_preference\_metrics`, and

  `main\_worklog\_path`.

- `global\_preferences`, especially when selected root preference

  metrics are empty.

- `worklog\_ground\_truth`, especially summarized before/after

  markdown changes.

- `rewritten\_message\_log`, only as supporting reference-trace

  evidence.

- `candidate\_a.agent\_trace`.

- `candidate\_b.agent\_trace`.

- Trace/message-log content may be head-tail truncated when `truncated` is true.

First decide whether the GT change updates the explicit `main\_worklog\_path`.

If it does, writing the durable main worklog is an important preference signal.

If it does not, exact artifact filename is much less important and content/style

alignment should dominate.

Comparison rubric:

1. Durable record / target behavior

   - Prefer the candidate that updates the same durable record type as GT.

   - If GT updates the explicit main worklog, prefer the candidate that writes

     that main worklog. A candidate that only writes a standalone artifact can

     still get content credit, but should usually lose this dimension.

   - If both write or both miss the main worklog, compare the quality of the

     written record rather than double-counting path.

2. Worklog language style

   - Prefer concise rough worklog notes when GT is concise.

   - Prefer mixed Chinese-English technical note-taking, short bullets,

     lightweight status wording, and local worklog rhythm when GT uses them.

   - Penalize converting a short daily/worklog note into a formal report,

     architecture spec, or over-polished deliverable unless GT itself is formal.

3. Judgment/status/uncertainty style

   - Prefer candidates that preserve tentative wording, TODO/pending/running

     status, blocked/failed checks, "still checking" language, rough comparative

     judgments, and non-overconfident conclusions when GT uses them.

   - Penalize unsupported completion claims, over-certainty, or hiding

     limitations in a polished artifact.

4. Proportional detail and scope control

   - Prefer the candidate whose amount of detail matches the reference change.

   - If GT is a short worklog continuation, a very long standalone document may

     be worse even if technically useful.

   - If GT is an artifact/spec, a longer artifact may be fine, but added details

     should not invent requirements or obscure the core preference.

5. Root metric support

   - Use selected root preference metrics as high-level constraints. Prefer the

     candidate that better supports those metrics in a way visible from the

     message log.

   - If selected root metrics are empty, use global person-level preferences

     instead of blocking.

   - Explain which selected root or global preference dimensions actually

     separate the two candidates.

Do not judge by:

- Which candidate has more tokens.

- Which candidate has a more formal file.

- Which candidate sounds more complete in final claims.

- Exact path/name matching for non-primary artifact documents.

- Exact URL/API/command equality when the concrete-recording style is otherwise

  preserved.

Suggested dimensions:

- `durable\_record\_target`

- `worklog\_concision\_and\_tone`

- `status\_uncertainty\_and\_judgment`

- `proportional\_scope`

- `root\_metric\_alignment`

Each dimension should choose one of:

`candidate\_a|candidate\_b`.

Decision labels:

- `candidate\_a`: candidate A is clearly better.

- `candidate\_b`: candidate B is clearly better.

- Always choose either `candidate\_a` or `candidate\_b`. If both are weak, choose

  the less weak one.

- If the difference is small, still choose the slightly better candidate and use

  lower confidence.

Confidence:

- Use high confidence only when several dimensions point the same way.

- Use medium confidence when the winner is mostly from one major factor such as

  main-worklog target.

- Use low confidence when both candidates are similar, both fail the main

  preference, or message-log evidence is thin.

Return exactly this JSON object:

\{

  "module": "preference\_recall\_compare",

  "winner": "candidate\_a|candidate\_b",

  "confidence": 0.0,

  "dimension\_results": [

    \{"dimension": "durable\_record\_target", "winner": "candidate\_a|candidate\_b", "reason": "short concrete reason"\},

    \{"dimension": "worklog\_concision\_and\_tone", "winner": "candidate\_a|candidate\_b", "reason": "short concrete reason"\},

    \{"dimension": "status\_uncertainty\_and\_judgment", "winner": "candidate\_a|candidate\_b", "reason": "short concrete reason"\},

    \{"dimension": "proportional\_scope", "winner": "candidate\_a|candidate\_b", "reason": "short concrete reason"\},

    \{"dimension": "root\_metric\_alignment", "winner": "candidate\_a|candidate\_b", "reason": "short concrete reason"\}

  ],

  "summary": "short concrete explanation of why the winner is better"

\}
\end{promptbox}

\subsubsection{Reference Historical Evidence}

For each target task, the following prompt is applied independently to every earlier task in the same participant workflow. Relevant pairs and excerpts extracted from the earlier trace form the reference historical evidence.

\begin{promptbox}{Pairwise Historical Relevance and Excerpt Extraction}
You are determining whether two Membench subtasks have prerequisite-memory relevance.

Each request contains one pair:

- `previous\_task`: the earlier subtask.

- `target\_task`: the subtask being evaluated.

- Both tasks include a task card and the original execution message log.

Objective:

Determine whether `previous\_task` provides useful prior memory for `target\_task`. Useful does not mean broadly similar. It means that remembering the earlier task would make the agent's search path, implementation choice, validation method, task boundary, output document, risk judgment, or final deliverable more accurate.

Use a strict standard:

- Default to `not relevant`. Mark the pair relevant only when the earlier task contains concrete information that would change execution or output for the target task.

- Relevant evidence must establish three links: the concrete fact in `previous\_task`, why `target\_task` needs it, and how it changes current execution, validation, documentation, or conclusion boundaries.

- If the evidence says only that the tasks share a project, worklog, file, time period, or topic, or that historical content should be preserved, normally mark it not relevant.

- Knowing only the primary-worklog path, document title, or requirement not to overwrite existing content is generally global document state rather than strong task-specific prerequisite memory.

- Empty placeholders, path mappings, ordinary formatting continuity, simple date appends, and avoidance of overwriting old content are normally not relevant. The exception is when recovering, locating, or repairing that exact document state is the target task's core objective and the previous task is unique or essential evidence.

- Consecutive updates to the same worklog are not automatically relevant. Relevance requires that the target task need a concrete conclusion, TODO, failure cause, experimental result, artifact, command, dataset, schema, code path, or judgment boundary from the earlier task.

- Broad conceptual analogy is insufficient. Statements such as both involve agents, environments, terminals, benchmarks, or failure analysis must identify the exact earlier fact that reduces wasted work or improves the current deliverable.

Relevant cases include:

- The target clearly continues the same pipeline, dataset, repository, experiment, document, prompt, evaluation, cleaning process, deployment process, or analysis thread.

- The previous task created or modified an artifact, schema, script, configuration, command, path, field, sample, conclusion, failure cause, TODO, or validation method needed by the target.

- A mainline worklog and a non-mainline artifact, report, specification, or prompt may be relevant to each other when the connection changes how the target task should be performed.

Not-relevant cases include:

- The tasks merely concern the same participant or broad project, occur near each other, or both involve documentation.

- They merely use the same primary worklog, file path, title, date format, or indentation style.

- The previous task wrote old content that a later edit to the same file must preserve.

- The previous task only created an empty file, placeholder, directory, title, or path mapping.

- The connection is only generic background, broad style, personality, formatting cleanup, or a content-free placeholder.

- The tasks share a broad topic but no concrete artifact, decision, data, validation, task boundary, or reusable conclusion.

- Remembering the previous task would not make the target faster, more accurate, less wasteful, or closer to the required deliverable.

Procedure:

1. Read both task cards to understand each objective and deliverable.

2. Read both message logs to determine what the agent actually inspected, ran, changed, verified, and summarized.

3. Ask whether remembering `previous\_task` would materially change the approach, eliminate irrelevant search, enable reuse of an existing artifact, avoid repeated validation, preserve an important boundary, or produce a more accurate conclusion.

4. If the answer cannot be stated beyond shared file, path, topic, historical preservation, or formatting continuity, mark the pair not relevant.

5. If relevant, explicitly reason through: concrete fact from the previous task -\textgreater{} concrete need in the target task -\textgreater{} effect on execution or output.

6. Write one concise `evidence` statement describing which fact from `previous\_task` affects `target\_task`.

7. In `memory\_needed`, state the essential information that must be remembered. Keep it short but understandable; do not repeat background.

8. If relevant, extract supporting excerpts from the original `previous\_task.message\_log`.

9. If not relevant, set `evidence` and `memory\_needed` to empty strings and `excerpts` to an empty array.

Excerpt requirements:

- Extract excerpts only from `previous\_task.message\_log`, never from `target\_task`.

- Preserve the source wording as closely as possible. Do not rewrite it as your own summary or invent facts absent from the previous task.

- Each excerpt should be a compact, evidence-dense passage, usually about 100--300 Chinese characters or a comparable amount of English text. It may combine consecutive or adjacent sentences from the same local context, but must not copy a large portion of the log.

- Prefer concrete evidence: filenames, paths, artifact names, run IDs, checkpoints, model names, dataset names, API endpoints, commands, scripts, metric values, sample counts, distributions, error messages, blockers, failure causes, user preferences, output formats, prohibited actions, completion state, and next steps.

- Do not extract vague statements such as "the user is working on a project," "this concerns a benchmark," or "further analysis is needed."

- A relevant pair normally contains 3--5 excerpts.

- Each excerpt's `reason` must explain only why that source passage is useful to `target\_task`.

Return strict JSON only:

\{

  "relevant": true,

  "evidence": "concise judgment, or empty when not relevant",

  "memory\_needed": "essential information to remember, or empty when not relevant",

  "excerpts": [

    \{

      "text": "an evidence-dense excerpt copied from previous\_task.message\_log",

      "reason": "why this excerpt is useful to target\_task"

    \}

  ]

\}

Rules:

- `relevant` must be `true` or `false`.

- Do not output Markdown or extra fields.

- Keep `evidence` and `memory\_needed` empty when they are unnecessary and concise when present.

- `excerpts` must be an array and must be `[]` when not relevant.

- Each excerpt object may contain only `text` and `reason`.

- Do not mark a pair relevant merely because the tasks are close in time.

- Do not be overly conservative merely because the dependency is indirect, but require a concrete fact that changes target-task execution or output.

Runtime user message:

Determine whether `previous\_task` provides useful prior memory for `target\_task`. Input:

\{

  "person": "\textless{}person\_id\textgreater{}",

  "previous\_task": \{

    "task\_card": \textless{}previous\_task\_card\textgreater{},

    "message\_log": \textless{}previous\_agent\_trace\textgreater{}

  \},

  "target\_task": \{

    "task\_card": \textless{}target\_task\_card\textgreater{},

    "message\_log": \textless{}target\_agent\_trace\textgreater{}

  \}

\}
\end{promptbox}

\subsubsection{Memory Diagnostic Prompts}

\paragraph{Relevance.}
The embedding-based mean similarity, recall, and precision are computed directly and therefore do not use an LLM prompt. The following prompt produces the complementary 1--5 relevance score.

\begin{promptbox}{Memory Relevance Score}
You are a strict evaluator for the memory\_relevance dimension of one Membench memory ablation subtask.

Category: context.

Question: Were the recalled memories actually relevant prior context for the current task?

This dimension evaluates the recalled context visible in `with\_recall.message\_log`, not whether the agent used it well. When present, recalled memory is inside `\textless{}memory\_recall\_context\textgreater{}...\textless{}/memory\_recall\_context\textgreater{}` in an early user message. Use the surrounding task in the same message to judge relevance. Use `without\_recall.message\_log` only as task and trajectory context, not as recalled memory.

What to judge:

- Identify the current task intent and required background.

- Identify recalled prior tasks, artifacts, decisions, worklogs, conventions, or user constraints from the visible recalled memory block.

- Decide which recalled memories are relevant, partially relevant, unrelated, stale, or potentially misleading.

- Prefer task-specific historical continuity over generic workspace history. A prior task that merely says "inspect the workspace" is weak unless it names artifacts or decisions needed now.

- If recall includes many irrelevant memories but also one clearly important item, score based on both signal and noise.

Do not judge:

- Whether with\_recall executed efficiently.

- Whether without\_recall recovered the same context.

- General task quality.

Score on a 1-5 scale:

1. Recall is absent, almost entirely unrelated, or actively misleading for the current task.

2. Recall has weak/generic relevance; useful anchors are sparse or buried in noise.

3. Recall contains some relevant prior context, but important anchors are missing, vague, or mixed with substantial noise.

4. Recall contains clearly relevant prior artifacts/decisions/worklog context with manageable noise.

5. Recall is highly relevant and specific, naming the key prior tasks/artifacts/decisions needed for this subtask.

Return exactly this JSON object:

\{

  "dimension": "memory\_relevance",

  "category": "context",

  "score": "integer 1-5",

  "relevant\_prior\_tasks": [

    \{"id": "subtask\_xxxx or history id", "relevance": "low|medium|high", "reason": "brief reason", "anchors": ["artifact/decision/convention"]\}

  ],

  "irrelevant\_or\_noisy\_memories": ["brief item"],

  "missing\_expected\_context": ["brief item"],

  "evidence": ["brief concrete evidence from Evaluation input"],

  "confidence": 0.0,

  "summary": "short judgment"

\}
\end{promptbox}

\begin{promptbox}{Environment Familiarity Score}
You are a strict evaluator for the environment\_familiarity dimension of one Membench memory ablation subtask.

Category: ability.

Question: Did the with\_recall run behave more familiar with the local workspace/project environment than the without\_recall run?

This is the renamed and refined routing\_gain dimension. It measures local-environment familiarity shown through navigation, search, and target selection. It is not a pure recall-relevance score.

Important neutrality rule:

- Checking `/workspace/context`, listing the root workspace, or broad discovery at the start is often required by task instructions. Do not count those checks alone as evidence of familiarity or unfamiliarity.

- Count only differences beyond shared setup behavior: specific historical file names, targeted directories, known artifact names, project terms, or reduced irrelevant detours.

What to judge:

- Whether with\_recall uses more specific workspace/project queries earlier.

- Whether it reaches relevant files, folders, prior artifacts, or project state with fewer irrelevant detours.

- Whether without\_recall has to reconstruct context through broad scans, wrong files, or avoidable command/path mistakes.

- Whether with\_recall knows where outputs should go in the local environment.

- Ignore failures clearly due to network outage, unavailable external services, or shared tool instability rather than agent choice.

- Use tool calls, assistant messages, and file/content references in `with\_recall.message\_log` and `without\_recall.message\_log`.

Score on a 1-5 scale:

1. With\_recall is less familiar: follows wrong local context, misses obvious files, or has more avoidable local detours than without\_recall.

2. Little environment familiarity advantage; both mostly rely on broad scans or with\_recall is only slightly more targeted.

3. Some familiarity: with\_recall uses a few relevant local anchors earlier, but without\_recall recovers them with similar effort.

4. Clear familiarity: with\_recall navigates to relevant project files/artifacts more directly and avoids notable detours.

5. Strong familiarity: with\_recall immediately or near-immediately uses precise local anchors/conventions and without\_recall shows clear reconstruction cost.

Return exactly this JSON object:

\{

  "dimension": "environment\_familiarity",

  "category": "ability",

  "score": "integer 1-5",

  "with\_recall\_behavior": "brief concrete trajectory summary",

  "without\_recall\_behavior": "brief concrete trajectory summary",

  "familiarity\_signals": ["specific signal"],

  "neutral\_setup\_ignored": ["generic setup behavior ignored"],

  "avoidable\_detours\_or\_failures": ["detour/failure and variant"],

  "evidence": ["brief concrete evidence from Evaluation input"],

  "confidence": 0.0,

  "summary": "short judgment"

\}
\end{promptbox}

\begin{promptbox}{Tool-call Purpose Classification}
You label the purpose of every tool call in an agent execution trace.

You receive the task instruction, the complete execution trace after limited tool-output truncation, and the exact call IDs that must be labeled. Classify each tool call using its purpose in context, not merely its tool name.

Use exactly one of these labels:

- `exploration`: The call is used to understand or recover the existing environment, workspace, data, dependencies, or prior state before deciding how to act. Examples include listing directories, searching for files or symbols, reading existing files, inspecting configuration, checking installed tools, and probing the current state. A targeted read can still be exploration when its purpose is to locate or understand the problem.

- `execution`: The call directly advances the requested task. This includes editing or creating deliverables, installing a required dependency, running the target computation, generating data or artifacts, applying a fix, and testing or validating the implementation or output. Failed calls remain `execution` when their intended purpose was to perform or validate the task.

- `other`: The call neither explores the environment nor executes the task itself. Typical examples are plan/status bookkeeping, user communication, or unrelated administrative actions.

Important distinctions:

- Do not label by command name alone. Reading a pre-existing file to understand the workspace is usually `exploration`; reading a newly generated output to verify the task is usually `execution`.

- Broad discovery and targeted diagnosis are both `exploration` when they primarily gather information needed to choose the next action.

- Tests, builds, smoke checks, and result inspection are `execution` when they validate work performed for the current task.

- A tool call that attempts the task but fails is still `execution`; the outcome does not change its purpose.

- Label every occurrence in the provided call-ID list exactly once and preserve its order. A call ID may appear more than once when the source trace reuses an ID; in that case, return the same ID once for each occurrence. Do not invent, omit, merge, or reorder occurrences.

Return only a JSON array in the same order as the provided call IDs:

[

  \{

    "call\_id": "exact call id",

    "label": "exploration"

  \}

]

Each object must contain exactly `call\_id` and `label`. Do not include explanations or Markdown.
\end{promptbox}

\begin{promptbox}{Execution Problem Identification}
You are evaluating an agent execution trace. Identify only concrete execution problems that the agent actually encountered while carrying out the task, and determine whether each problem was resolved later in the same trace.

Count problems such as:

- a command, tool call, test, build, or program that failed unexpectedly;

- a missing local file, path, dependency, or executable that blocked an intended operation;

- an invalid local configuration, schema, data format, or implementation exposed during execution.

Do not count:

- weaknesses in the final answer, incomplete content coverage, stylistic issues, uncertainty, risks, plans, or suggestions;

- normal exploration, an expected negative check, or a search that legitimately returns no matches;

- a possible problem that was only discussed but never observed;

- failures caused by an unavailable external service, network, rate limit, authentication, credentials, permissions outside the workspace, or another resource the agent cannot repair.

Merge repeated failures caused by the same underlying issue. Keep separate problems with different causes.

Set "resolved" to true only when later execution evidence shows that the problem was fixed, or that a lossless alternative completed the same intended operation. The agent merely saying that it fixed the issue is not sufficient. Skipping the operation, accepting reduced functionality, leaving a TODO, or using a workaround that loses part of the requested result is unresolved.

Return only a JSON array. Each item must contain exactly:

\{"problem": "a concise description of the observed execution problem", "resolved": true\}

Use false instead of true when unresolved. Return [] when no countable execution problem occurred.
\end{promptbox}

\begin{promptbox}{Problem Solvability from Recall}
You are given candidate memory, a without-memory execution trace, and a fixed list of execution problems that remained unresolved in that trace.

For each problem, decide whether the candidate memory contains enough concrete information to solve it.

Set memory\_can\_solve to true only when the candidate memory provides a specific fact, correct path, constraint, existing artifact, verified command, configuration, or known procedure that addresses the actual cause of the problem. If an agent had this information and used it correctly, it should be able to resolve the problem without losing part of the requested result.

Set memory\_can\_solve to false when the memory is only topically related, provides general background, repeats the task goal, or lacks the concrete information needed to resolve the problem. Do not assume that information absent from the supplied memory was available elsewhere.

Do not identify new problems, rewrite the supplied problems, or reassess whether they were resolved.

Return only a JSON array in the same order as the fixed problems. Each item must contain exactly:

\{"problem\_index": 0, "memory\_can\_solve": false\}

Include every supplied problem index exactly once. Return [] when the fixed problem list is empty.
\end{promptbox}

\begin{promptbox}{Hallucination Robustness Score}
You are a strict evaluator for the hallucination\_risk dimension of one Membench memory ablation subtask.

Category: context.

Question: Did recalled memory cause or risk hallucinated, stale, irrelevant, or overfit behavior?

This replaces misuse\_risk with a clearer direction: higher score means lower hallucination/misuse risk and better grounding.

Judge whether with\_recall:

- Grounds historical claims in actual recalled or workspace-visible evidence.

- Avoids inventing prior decisions, files, constraints, or user preferences.

- Avoids following irrelevant recalled memory into the wrong artifact/task/person.

- Correctly treats weak or irrelevant recall as weak instead of overusing it.

- Does not ignore current-task instructions because of old memory.

- Judge only memory-related risk visible in `with\_recall.message\_log`. If no recalled memory is visible, do not invent a memory misuse case.

Hallucination/misuse examples:

- Claims an artifact or convention exists without evidence.

- Updates a file because of stale/irrelevant memory rather than current task needs.

- Imports another person's or another project's context.

- Overfits to a previous task and solves the wrong problem.

- Treats a generic old workspace scan as a task-specific constraint.

Score on a 1-5 scale, where higher is safer:

1. Severe memory-induced hallucination or misuse; with\_recall is materially harmed.

2. Clear risk or some harmful overfitting to irrelevant/stale memory.

3. Some uncertainty/noise; with\_recall mostly avoids harm but makes weakly grounded historical assumptions.

4. Low risk; with\_recall is mostly grounded and ignores irrelevant recall appropriately.

5. Very low risk; with\_recall carefully grounds memory use, verifies important claims, and avoids overuse of weak recall.

Return exactly this JSON object:

\{

  "dimension": "hallucination\_risk",

  "category": "context",

  "score": "integer 1-5",

  "risk\_level": "low|medium|high",

  "grounded\_memory\_use": ["specific grounded behavior"],

  "hallucination\_or\_misuse\_cases": ["specific case if any"],

  "irrelevant\_recall\_handling": "brief judgment",

  "evidence": ["brief concrete evidence from Evaluation input"],

  "confidence": 0.0,

  "summary": "short judgment"

\}
\end{promptbox}

\begin{promptbox}{Memory-induced Problem Classification}
You are given a task instruction, the memory recalled before execution, an execution trace, and a fixed list of execution problems observed in that trace.

For each fixed problem, decide whether the recalled memory caused or materially contributed to the problem.

Mark memory\_induced as true only when all of the following are supported by the inputs:

1. The recalled memory contains incorrect, outdated, misleading, or ambiguous information relevant to the problem.

2. The execution trace shows that the agent actually relied on or adopted that information.

3. This reliance caused or materially contributed to the observed problem.

Mark it as false when the memory merely contains questionable information that the agent did not use, when the connection is only speculative, or when the problem is a general coding, command, dependency, reasoning, or execution error not caused by recalled memory.

Do not identify new problems and do not change the supplied problem descriptions or resolution labels.

Return only a JSON array in the same order as the fixed problems. Each item must contain exactly:

\{"problem\_index": 0, "memory\_induced": false\}

Include every supplied problem index exactly once. Return [] when the fixed problem list is empty.
\end{promptbox}

\end{document}